\documentclass[format=acmsmall, review=false, screen=true]{acmart}
\usepackage{booktabs} 
\usepackage{color}
\usepackage{subfigure} 
\usepackage{array}
\usepackage{breqn}
\usepackage{url}
\usepackage{balance}
\usepackage{multirow,tabularx}
\usepackage{balance}
\usepackage{supertabular}
\usepackage{booktabs,longtable}
\usepackage{hhline}
\usepackage[flushleft]{threeparttable}
\usepackage{algorithm}
\usepackage{algorithmic}
\usepackage{epsfig}
\usepackage{graphicx}
\usepackage{epstopdf}
\usepackage{thmtools}
\usepackage{enumitem}
\usepackage{verbatim}
\usepackage{amsfonts}
\usepackage{hyperref}
\hypersetup{hidelinks,
colorlinks=false} 
\usepackage{makecell}
\AtBeginDocument{%
  \providecommand\BibTeX{{%
    \normalfont B\kern-0.5em{\scshape i\kern-0.25em b}\kern-0.8em\TeX}}}

\copyrightyear{2023}
\acmYear{2023}
\setcopyright{acmcopyright}
\acmDOI{10.1145/3556537}
\acmISBN{978-1-4503-6317-4/19/07}
\acmJournal{CSUR}
\acmArticle{188}

\begin{document}

\title{A Survey on Video Moment Localization}
\titlenote{Several mistakes are corrected from the published version. Liqiang Nie is the corresponding author. }

\author{Meng Liu}
\email{mengliu.sdu@gmail.com}
\affiliation{
  \institution{Shandong Jianzhu University}
  \country{China}
  }
\author{Liqiang Nie}
\email{nieliqiang@gmail.com}
\affiliation{
  \institution{Harbin
Institute of Technology (Shenzhen)}
  \country{China}
  }
\author{Yunxiao Wang}
\email{yunxiao.wang@mail.sdu.edu.cn}
\affiliation{
  \institution{Shandong University}
  \country{China}
  }
\author{Meng Wang}
\email{eric.mengwang@gmail.com}
\affiliation{
  \institution{Hefei University of Technology}
  \country{China}
  }
\author{Yong Rui}
\email{yongrui@lenovo.com}
\affiliation{
  \institution{Lenovo Company Ltd.}
  \country{China}
  }

\begin{abstract}
 Video moment localization, also known as video moment retrieval, aiming to search a target segment within a video described by a given natural language query. Beyond the task of temporal action localization whereby the target actions are pre-defined, video moment retrieval can query arbitrary complex activities. In this survey paper, we aim to present a comprehensive review of existing video moment localization techniques, including supervised, weakly supervised, and unsupervised ones. We also review the datasets available for video moment localization and group results of related work. In addition, we discuss promising future directions for this field, in particular large-scale datasets and interpretable video moment localization models. 
\end{abstract}


\maketitle

\section{Introduction}
\begin{figure}[t]
\centering
  \includegraphics[width=0.9\textwidth]{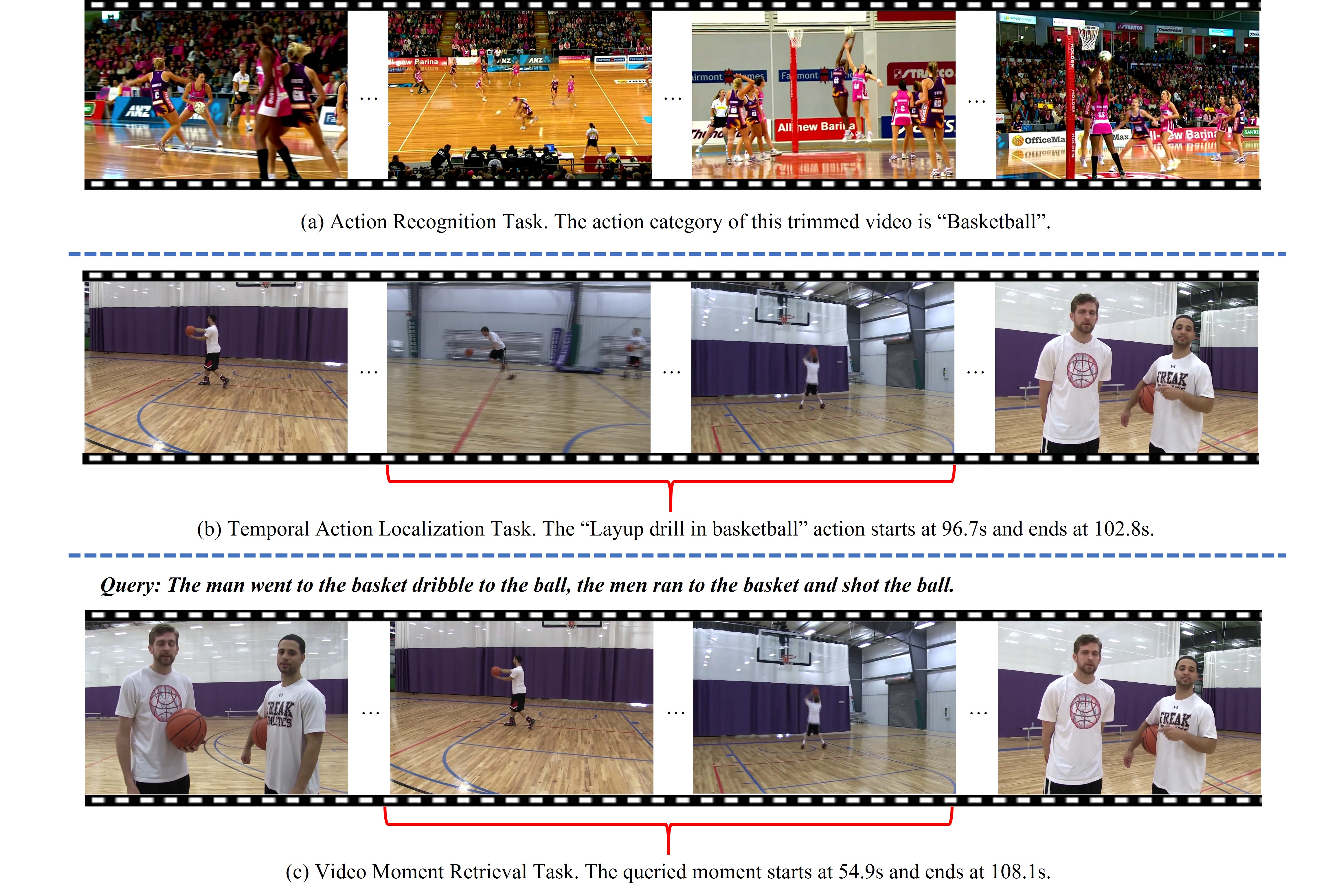}
  \vspace{-3ex}
\caption{Illustration of the action recognition, temporal action localization, and video moment localization (retrieval) tasks. (a) The action recognition task aims to classify trimmed videos into different action categories. (b) The goal of temporal action localization is to automatically recognize the interval of the action occurs from the given untrimmed video and judge the category of the detected action. And (c) the purpose of the video moment localization is to localize the temporal moment referenced in the query.}\label{fig1}
\vspace{-2ex}
\end{figure}

With the increasing prevalence of digital cameras and social networks, plenty of videos are recorded, stored, and shared daily~\cite{micro-video}, especially the surveillance videos. Such large-scale video data prompts video content analysis to become increasingly essential~\cite{9211791}. Compared with static images, one critical characteristic of videos is that they could depict the behavior evolution of a certain object over time. Therefore, understanding and recognizing actions has become the fundamental task to effectively analyze videos. In the past decade, a large body of literature has paid attention to action recognition in videos~\cite{8600333}. And the vast majority of existing studies tackle the action recognition problem as the video classification task~\cite{classification}, where the trimmed videos contain a single action (as shown in Fig.~\ref{fig1}(a)). However, in real-world applications, such as surveillance~\cite{surveillance}, robotics~\cite{robot}, and autonomous driving~\cite{driving}, cameras continuously record video streams. In other words, videos, in reality, are mostly untrimmed. Therefore, developing algorithms that simultaneously decide both the temporal intervals and the action categories as they occur is indispensable. 

Inspired by this, temporal action localization (as illustrated in Fig.~\ref{fig1}(b)), which jointly classifies action instances and localizes them in an untrimmed video, becomes a vital task in video understanding~\cite{9171561}. But existing temporal action localization methods are restricted to a pre-defined list of actions. They are hence inflexible since activities in the wild are even more complex, like ``The man went to the basket dribble to the ball, the men ran to the basket and shot the ball''. Considering the natural language expression could vary according to the content of videos, it is more natural to utilize the natural language query to localize the desired moment from the given video, as shown in Fig.~\ref{fig1}(c). Accordingly, video moment localization, aiming to identify the specific start and end timestamps of the moment in response to the given query, has been a hot research topic recently. 
 
While this task opens up great opportunities to better video understanding, it is substantially more challenging due to the following reasons: 
1) The given queries can be arbitrarily complex natural descriptions. Considering the query ``The black cat jumps back up after falling'' as an example, it describes the temporal relationship between ``jump back up'' and ``fall'' actions in a video. Moreover, the language sequence is misaligned with video sequence. Therefore, how to well comprehend the complex query information is one barrier for video moment localization.
2) Different moments in the untrimmed video have varying durations and diverse spatial-temporal characteristics. 
In addition, a long video often contains multiple moments of interest. For example, given a typical query ``A Ferris wheel second comes into view'', the model requires to find the second occurrence of ``Ferris wheel''. Thereby, how to effectively distinguish relevant video moments and precisely localize the moment of interest is very difficult.   
3) As a multimodal task, how to model the complex correlations between videos and queries is crucial. 
And 4) how to boost the efficiency of the localization method is an important issue that matters practical use. 

\begin{figure}[t]
	\centering
  \subfigure[supervised video moment localization]{
  \label{fig:sup}\includegraphics[width=0.8\textwidth]{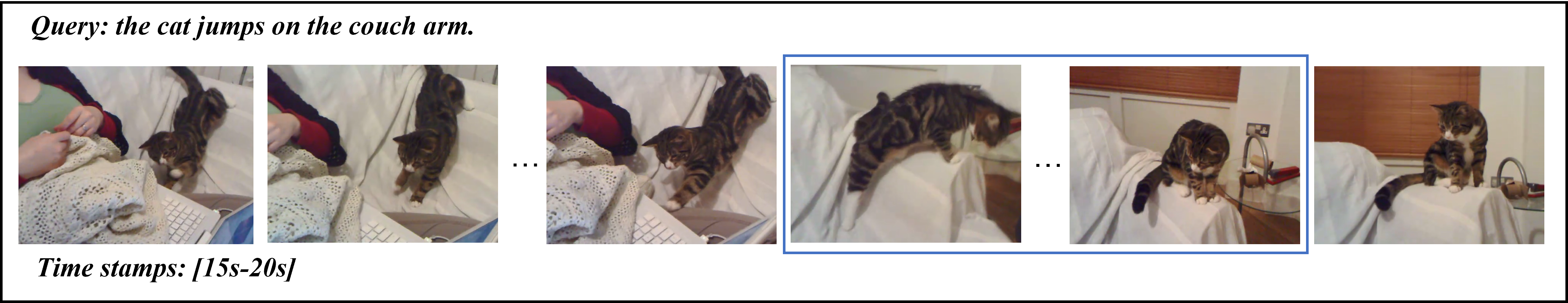}\vspace{-4ex}}
  \subfigure[weakly supervised video moment localization]{
  \label{fig:weak}\includegraphics[width=0.8\textwidth]{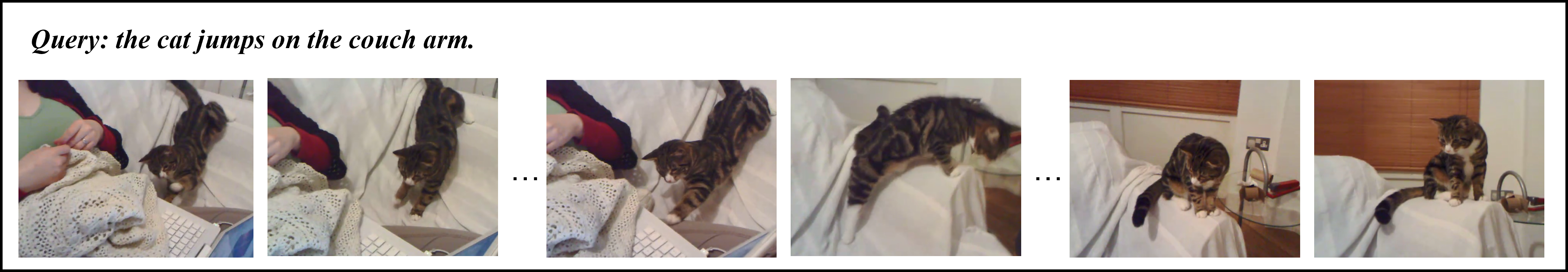}\vspace{-4ex}}
    \vspace{-4ex}
\caption{Visualization of the difference between supervised and weakly supervised video moment localization.}\label{diff}
\vspace{-2ex}
\end{figure}

Recently, deep learning techniques have emerged as powerful methods for various tasks, such as video captioning~\cite{video-cap} and text-video retrieval~\cite{video-text1}. This motivates many researchers to resort to deep learning approaches to tackle the video moment localization task. 
Particularly, existing work can be divided into three categories: 1) deep supervised learning approaches, 2) deep weakly supervised learning ones, and 3) deep unsupervised learning ones. The main difference between the supervised and weakly supervised learning methods is the availability of the exact temporal annotation of the moment corresponding to the given query, as shown in Fig.~\ref{diff}. To be specific, the former one 
requires the full annotations of temporal boundaries for each video and the given query, such as video segment of (15s,20s) in Fig.~\ref{fig:sup}. In contrast, the latter one merely needs video-sentence pairs for training, and they do not require the temporal boundary annotation information. Different from the aforementioned two categories, unsupervised video moment localization methods require no textual annotations of videos, namely they only require easily available text corpora and a collection of videos to localize.

\begin{table}[t]
\caption{Summarization of supervised video moment localization approaches.}
\vspace{-2ex}
    \centering
   \scalebox{0.6}{ \begin{tabular}{|c|c|ccc|c|}
   \hline
Architecture & Moment Generation& Method & Published&  Year& Key Aspect \\
    \hline

\multirow{15}{*}{Two-stage}&\multirow{3}{*}{\shortstack{Hand-crafted \\ Heuristics}} &  MCN~\cite{Lisa2017Localizing}&ICCV&2017& Pioneer Work\\
\cline{3-6}
&&MLLC~\cite{Lisa2018Localizing}&EMNLP&2018&Visual Modeling\\
\cline{3-6}
&&TCMN~\cite{Songyang2019Exploiting}& ACM MM&2019&Query Modeling\\
\cline{2-6}
&\multirow{9}{*}{\shortstack{Multi-scale \\Sliding Windows}}&CTRL~\cite{Jiyang2017TALL}&ICCV&2017&Pioneer Work\\
\cline{3-6} 
&&ACRN~\cite{Meng2018Attentive}&SIGIR&2018& \multirow{4}{*}{Visual Modeling}\\
&&VAL~\cite{Xiaomeng2018VAL}&PCM&2018&\\
&&SLTA~\cite{Bin2019Cross}&ICMR&2019&\\
&&MMRG~\cite{mmrg}&CVPR&2021&\\
 \cline{3-6}
&&ROLE~\cite{liumm2018}&ACM MM&2018&Query Modeling\\
 \cline{3-6}
&&MCF~\cite{Aming2018Multi}& IJCAI&2018 &\multirow{2}{*}{Inter-modal Interaction Modeling}\\ 
&&ACL~\cite{Runzhou2019MAC}&WACV&2019&\\
\cline{3-6}
&&MIGCN~\cite{migcn}&IEEE TIP &2021&Inter- and Intra-modal Interaction Modeling \\
\cline{2-6}
&\multirow{3}{*}{\shortstack{Moment \\Generation Networks}}&QSPN~\cite{Huijuan2019Multilevel}&AAAI&2019&Query-guided Proposal  Generation\\
 
&&SAP~\cite{Shaoxiang2019Semantic}&AAAI&2019&Visual Concept-based Proposal Generation\\
&&BPNet~\cite{BPNet}&AAAI&2021&VSLNet~\cite{Hao2020Span} based Proposal Generation\\
\cline{1-6}
\multirow{47}{*}{One-stage}&\multirow{10}{*}{Anchor-based}&TGN~\cite{Jingyuan2018Temporally}&EMNLP&2018&Pioneer Work\\
\cline{3-6}
&&BSSTL~\cite{Cheng2019Bidirectional} &ICIP&2019& \multirow{4}{*}{Inter-modal Interaction Modeling} \\
&&RMN~\cite{liu-etal-2020-reasoning}& ACL COLING &2020  &\\
&&FIAN~\cite{Xiaoye2020Fine}& ACM MM&2020 &\\
&&I2N~\cite{iin}&IEEE TIP&2021&\\
\cline{3-6}
&&CMIN~\cite{Zhu2019Cross} &SIGIR&2019&\multirow{4}{*}{Inter- and Intra-modal Interaction Modeling} \\
&&CSMGAN~\cite{Daizong2020Jointly}& ACM MM&2020 & \\
&&PMI-LOC~\cite{Shaoxiang2020Learning}& ECCV&2020 & \\
&&CMIN-R~\cite{cmin_ext}& IEEE TIP& 2020&\\

 \cline{3-6}
&&CBP~\cite{Jingwen2020The}&AAAI&2020&Localization Module\\
\cline{2-6}
&\multirow{5}{*}{Temporal Convolution}&SCDM~\cite{Yitian2019Semantic}&NIPS&2019&Semantic Modulated Temporal Convolution\\
&&MAN~\cite{Da2019MAN}&CVPR&2019&Hierarchical Convolutional Network\\
&&CLEAR~\cite{clear}&IEEE TIP& 2021&Bi-directional Temporal Convolution\\
&&IA-Net~\cite{ia-net}&EMNLP& 2021&Inter- and Intra-modal Interaction Modeling\\
&&CMHN~\cite{cmhn}&IEEE TIP&2021&Efficiency\\
\cline{2-6}
&Segment-tree&CPN~\cite{cpn}&CVPR&2021&Segment-tree\\
\cline{2-6}
&\multirow{2}{*}{Proposal Generation}&APGN~\cite{APGN}&EMNLP&2021&Adaptive Proposal Generation Network\\
&&Lp-Net~\cite{LP-Net}&EMNLP&2021&Learnable Proposal Generation Network\\
\cline{2-6}
&\multirow{10}{*}{Enumeration}&TMN~\cite{Bingbin2018Temporal}&ECCV&2018&Pioneer Work\\
&&2D-TAN~\cite{Songyang2020Learning}&AAAI&2020&2D Temporal Map\\
&&MS-2D-TAN~\cite{MS-2D-TAN}&IEEE TPAMI&2021&Multi-scale 2D Temporal Map\\
\cline{3-6}
&&DPIN~\cite{DIN}&ACM MM&2020&\multirow{5}{*}{Interaction Modeling}\\
&&MATN~\cite{matn}&CVPR&2021&\\
&&SMIN~\cite{smin}&CVPR&2021&\\
&&SV-VMR~\cite{sv-vmr}&ICME&2021&\\
&&RaNet~\cite{ranet}&EMNLP&2021&\\
\cline{3-6}
&&DCM~\cite{dcm}&SIGIR&2021&Location bias\\
&&FVMR~\cite{fvmr}&ICCV&2021&Efficiency\\
\cline{2-6}
&\multirow{19}{*}{Proposal-free}&LNet~\cite{Jingyuan2019Localizing}&AAAI&2019&Pioneer Work \\
\cline{3-6}
&& ABLR~\cite{Yitian2019To}&AAAI&2019&\multirow{9}{*}{Inter-modal Interaction Modeling}\\
&&ExCL~\cite{Soham2019ExCL}&NAACL-HLT&2019& \\
&&SQAN~\cite{Jonghwan2020Local}&CVPR&2020&\\
&&PFGA~\cite{Cristian2020Proposal}&WACV&2020&\\
&&VSLNet~\cite{Hao2020Span}&ACL&2020& \\
&&VSLNet-L~\cite{VSLNet-L}&IEEE TPAMI&2021& \\
&&ACRM~\cite{acrm}& IEEE TMM&2021&\\
&&MIM~\cite{liang2021local}&ICMR&2021&\\
&&SSMN~\cite{SSMN}& ACM TOMM& 2021&\\
\cline{3-6}
&&HVTG~\cite{HTV}&ECCV&2020&\multirow{2}{*}{Inter- and Intra-modal Interaction Modeling}\\
&&MQEI~\cite{mqei}&IEE T-ITS&2021&\\
\cline{3-6}
&&CBLN~\cite{cbln}&CVPR&2021&\multirow{3}{*}{Visual Modeling}\\
&&CP-Net~\cite{cp-net}&AAAI&2021&\\
&&DORi~\cite{dori}&WACV&2021&\\
\cline{3-6}
&&DEBUG~\cite{Chujie2019DEBUG}&EMNLP&2019 & Imbalance  Problem\\
&&DRN~\cite{Runhao2020Dense}& CVPR&2020 & Dense Supervision \\
&&IVG~\cite{IVG}&CVPR&2021&Causal Interventions\\
&&DRFT~\cite{drft}&NIPS&2021&Multi-modal Information\\
\cline{1-6}
 
\multirow{7}{*}{Reinforcement}&\multirow{7}{*}{Proposal-free}& RWM~\cite{Dongliang2019Read}&AAAI&2019&Pioneer Work\\
\cline{3-6}
&& TSP-PRL~\cite{Jie2020Tree}&AAAI&2020&Policy\\

\cline{3-6}
&&SM-RL~\cite{Weining2019Language}&CVPR&2019&\multirow{3}{*}{Visual Modeling}\\
&& STRONG~\cite{strong}&ACM MM& 2020&\\
&& TripNet~\cite{Meera2019Tripping}&BMVC&2020&\\
\cline{3-6}
&& AVMR~\cite{AVMR}& ACM MM &2020&Reward\\
\cline{3-6}
&&MABAN~\cite{maban}&TIP&2021& Inter-modal Interaction Modeling\\
\hline
\end{tabular}}
    \label{tab1}
    \vspace{-2ex}
\end{table}

\begin{table}[t]
\caption{Summarization of weakly-supervised and unsupervised video moment localization approaches.}
\vspace{-2ex}
    \centering
   \scalebox{0.7}{ \begin{tabular}{|c|c|c|ccc|c|}
   \hline
Paradigm & Architecture & Moment Generation& Method & Published&  Year& Key Aspect \\
    \hline

\multirow{11}{*}{\shortstack{Weakly\\ Supervised}}&\multirow{4}{*}{Two-stage}&\multirow{4}{*}{\shortstack{Multi-scale \\ Sliding Windows}}&TGA~\cite{Niluthpol2019Weakly}&CVPR&2019&Pioneer Work\\
\cline{4-7}
&&& WSLLN~\cite{Mingfei2019Wslln}&EMNLP&2019& \multirow{2}{*}{Proposal Selection}\\
&&&VLANet~\cite{Minuk2020VLANet} &ECCV&2020 & \\
\cline{4-7}
&&&LoGAN~\cite{logan}&WACV&2021&Visual Modeling\\
\cline{2-7}
&\multirow{6}{*}{One-stage}& \multirow{2}{*}{Anchor-based}&SCN~\cite{Zhijie2020Weakly}&AAAI&2020&Scoring Refinement\\
&&&VCA~\cite{vca}&ACM MM&2021& Visual Modeling\\
\cline{3-7}
&&\multirow{4}{*}{Enumeration}&RTBPN~\cite{RTBPN}& ACM MM&2020& Intra-sample Confrontment\\
\cline{4-7}
&&&LCNet~\cite{lcnet}&TIP& 2021&Inter-modal Interaction Modeling\\
\cline{4-7}
&&&WSTAN~\cite{wstan}&TMM&2021&\multirow{2}{*}{Visual Modeling}\\
&&&MS-2D-RL~\cite{wstan}&ICPR&2021&\\
\cline{2-7}
&Reinforcement&Proposal-free&BAR~\cite{BAR}&ACM MM& 2020& Boundary Refinement\\
\cline{1-7}
\multirow{2}{*}{Unsupervised}&\multirow{2}{*}{One-stage}&Enumeration&U-VMR~\cite{u-vmr}&IEEE TCSVT&2021&Knowledge Distillation\\
\cline{4-7}
&&Proposal-free&PSVL~\cite{PSVL}&ICCV&2021&Pseudo-supervision Generation\\
\hline
\end{tabular}}
    \label{weak_m}
    \vspace{-2ex}
\end{table}

To give a comprehensive overview of this field, including models, datasets, and future directions, we summarize the work on video moment localization before 
Dec. 2021 and present this survey. Specifically, to perform a deeper analysis of existing approaches, we establish the fine-grained taxonomy of each category based on their architectures, moment generation strategies, and key characteristics. For instance, for deep supervised learning-based approaches (as summarized in Table~\ref{tab1}), we first group them into three categories: two-stage, one-stage, and reinforcement learning methods according to their architectures. Thereinto, two-stage methods commonly adopt separate schemes (e.g., the sliding windows) to generate moment candidates and then match them with the query sentences to obtain the target moment. One-stage ones integrate the moment generation and moment localization into a unified framework,  while reinforcement learning paradigms formulate the video moment localization task as a sequential decision making problem.
Afterwards, according to the strategies of moment generation, we respectively divide two-stage, one-stage, as well as reinforcement learning methods into different subclasses, and each subclass is further divided into several parts according to the characteristics of corresponding methods. Note that similar fine-grained taxonomies are established for weakly supervised and unsupervised learning methods, as reported in Table~\ref{weak_m}.

Compared to previous surveys on video moment localization~\cite{liu2021survey}\cite{yang2020survey}, the contributions of this survey are as follows: 
\begin{itemize}
\item Existing surveys pay more attention to analyzing supervised learning based localization methods, disregarding the analysis of weakly supervised and unsupervised ones. Differently, we provide a comprehensive survey of video moment localization approaches, including supervised, weakly supervised, and unsupervised learning based ones.
   \item We establish a holistic and fine-grained taxonomy for existing approaches according to their architectures, moment generation strategies, and key characteristics. This is advantageous to highlight the major strengths and shortcomings of existing methods. Nevertheless, other literature reviews mainly focus on organizing the typical approaches according to the coarse-grained taxonomy, such as localization policies~\cite{liu2021survey} or moment generation schemes~\cite{yang2020survey}.
    \item Our survey covers more papers on the video moment localization topic. To be specific, we summarize the work on video moment localization before Dec. 2021, while existing surveys mainly summarize the work published in 2019 and 2020. 
    \item  We introduce the methods of different categories in detail and conduct comparison among them. Nevertheless, these two surveys merely elaborate the typical methods of each category, as well as lack the deep analysis of other approaches.
\end{itemize}

The remainder of this paper is organized as follows. Section 2 briefly introduces some related research area. Section 3, 4, and 5 review supervised learning, weakly supervised learning, and unsupervised learning approaches on video moment localization, respectively. 
Section 6 describes the currently available datasets and evaluation metrics used for video moment localization. We separately analyze experimental results of existing approaches and discuss possible future research directions in Section 7 and 8.  We conclude the work in Section 9. 

\section{Related Research Area}
There are several research fields closely related to video
moment localization, which we now briefly describe.
\subsection{Temporal Action Localization}
Temporal action localization~\cite{su2021bsn++}\cite{qing2021temporal}, aiming at jointly classifying a pre-defined list of action instances and localizing the timestamps of them in an untrimmed video, is most relevant to the video moment localization task. The main difference between them is that video moment localization aims to localize the specific moment within the untrimmed video via the given natural language query. Apparently, these two tasks encounter some common
challenges, e.g., temporal proposal generation and temporal proposal relation modeling. Therefore, early video moment localization approaches directly apply temporal action localization techniques by merely replacing the action classification with cross-modal matching module. Recently, some video moment localization systems explore different grained interactions between the query and the moment as well as the query modeling to further increase both the accuracy and efficiency. Still, temporal action localization techniques will continue to serve as the foundation for the advance of its counterpart in the video moment localization.

\subsection{Video Object Grounding}
Existing video object grounding task can be broken down into the following categories: 1) localizing individual objects mentioned in the language query~\cite{ZhLoCoBMVC18}; 2) localizing a spatio-temporal tube of the target object based on the given language query~\cite{chen2019weakly} or person tracklet (i.e., video re-identification~\cite{li2020multi}\cite{li2019global}); and 3) only localizing  the referred objects in the language query\cite{sadhu2020video}. The main difference between the first and third category is that the former treats each query word independently and does not distinguish between different instances
of the same object, while the latter requires additional disambiguation using object-object relations in both time and space. Compared with the second subtask that requires both spatial and temporal localization, i.e., localizing a sequence of bounding
boxes of the queried object, video moment localization merely focuses on determining the temporal boundaries of events corresponding to the given sentence. There are many
key technical challenges, such as cross-modality modeling and fine-grained reasoning, that are shared between video object grounding and video moment localization.

\subsection{Video Corpus Moment Retrieval}
Video corpus moment retrieval aims to retrieve a short fraction in a video that semantically
corresponds to a text query, from a corpus of untrimmed
and unsegmented videos~\cite{zhang2021video}. Differently, video moment localization is to retrieve a short fraction from a single given video that corresponds to a text query. In fact, the task of video corpus moment retrieval was
extended from video moment localization by \cite{vcmr}, which better matches
real-world application scenarios. However, video corpus moment retrieval imposes the additional requirement to distinguish moments from different videos as compared to the task of video moment localization.

\section{Supervised Video Moment Localization}
\textbf{Problem Formulation.} We denote a video as $V = \{f_t\}_{t=1}^{T}$, where $T$ is the frame number of the video. Each video is associated with temporal annotations $A=\{(s_j,\tau_j^s,\tau_j^e)\}_{j=1}^M$, where $M$ is the annotation number of the video $V$ and $s_j$ is a sentence query with respect to a video moment that has $\tau_j^s$ and $\tau_j^e$ as start and end time points in the video. In the supervised video moment localization setting, the training data are the annotated query and video pairs (i.e., ($V$, $A$)). During the inference, given a video and a natural language query, the trained model would output the location information $(\tau^s,\tau^e)$ of the target moment regarding a given query. 

We summarize existing supervised video moment localization approaches in Table~\ref{tab1}. Specifically, they can be roughly grouped into three categories: two-stage, one-stage, and reinforcement learning-based approaches. In what follows, we detail them accordingly.

\subsection{Two-stage Methods}
\begin{figure}
    \centering
    \includegraphics[width=0.75\textwidth]{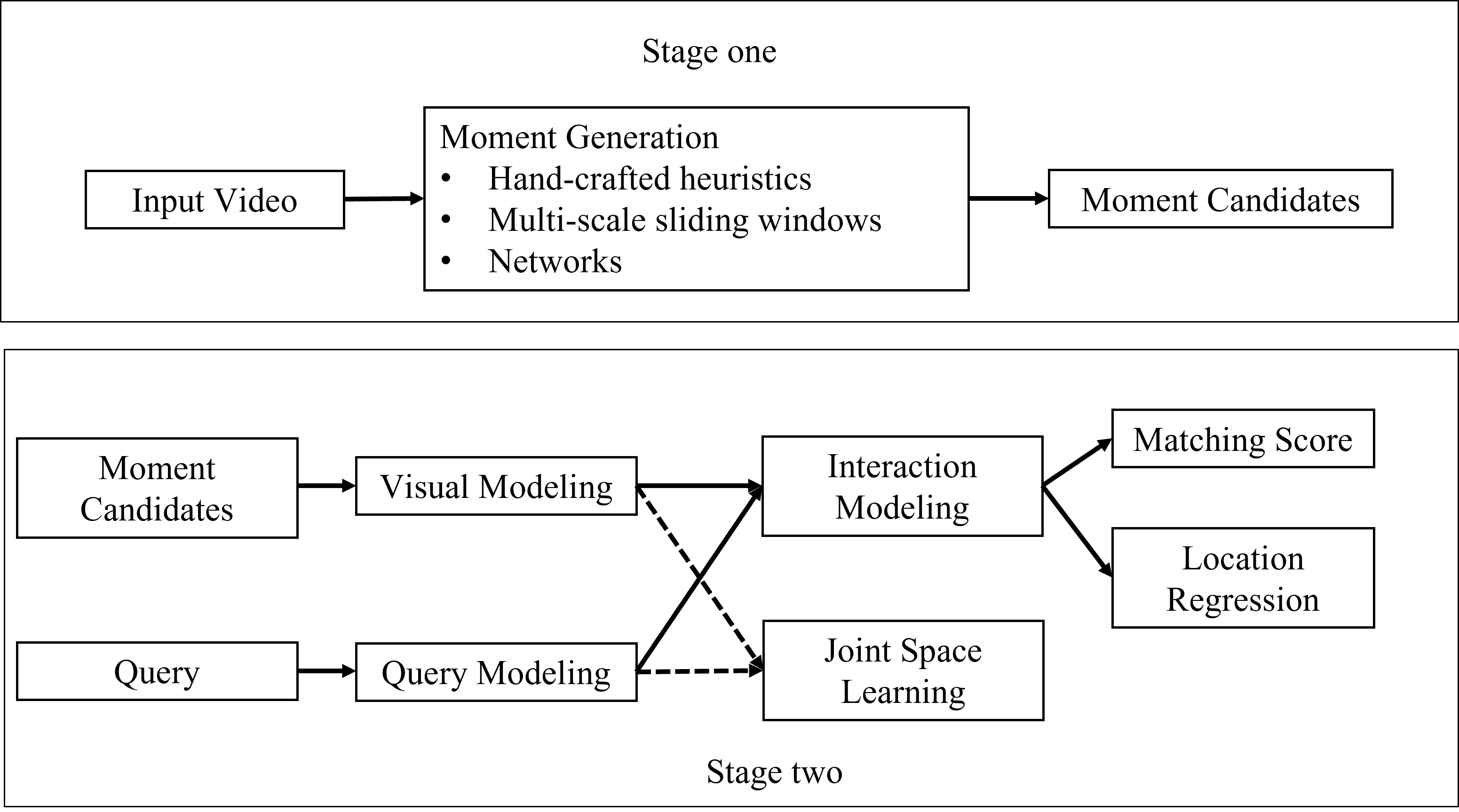}
    \vspace{-2ex}
    \caption{The schema of the two-stage methods in the whole paper.}
    \label{fig:two-stage}
\end{figure}

The two-stage supervised video moment localization approaches commonly utilize a separate scheme (e.g., the sliding windows) to generate moment candidates, and then match them with the query to find the target moment. A general diagram of two-stage video moment localization methods is shown in Fig.~\ref{fig:two-stage}. In particular, according to the manner of generating moment candidates, existing two-stage methods can be divided into three groups: hand-crafted heuristics, multi-scale sliding windows, and moment generation module based models~\cite{Huijuan2019Multilevel}. 

\subsubsection{\textbf{Hand-crafted heuristics}} 
The work of Hendricks et al.~\cite{Lisa2017Localizing} is generally regarded as the pioneer in this branch. It treats the video moment localization task as the moment retrieval problem and presents Moment Context Network (MCN)\footnote{\url{https://github.com/LisaAnne/LocalizingMoments}.}. To train and evaluate the proposed model,~\cite{Lisa2017Localizing} collects the Distinct Describable Moments (DiDeMo) dataset, which consists of over 40,000 pairs of natural language queries and localized moments in unedited videos. According to the characteristic of this dataset, \cite{Lisa2017Localizing} generates moment candidates via hand-crafted heuristics. Concretely, a video is first segmented into six five-second clips, and then any contiguous set of clips are utilized to construct a possible moment. Having obtained 21 moment candidates, it learns a shared embedding space for both video temporal context features and natural language queries, adopting the following inter-intra video ranking loss,
\begin{equation}
\begin{cases}
&L(\theta) = \lambda \sum_i L_i^{intra}(\theta)+(1-\lambda)\sum_iL_i^{inter}(\theta),\\
&L_i^{intra}(\theta)=\sum_{m_j\neq m_i}L^R(D_{\theta}(s_i,m_i),D_{\theta}(s_i,m_j)),\\
&L_i^{inter}(\theta)=\sum_{j\neq i}L^R(D_{\theta}(s_i,m_i),D_{\theta}(s_i,m_j)),
\end{cases}
\end{equation}
where $L_i^{intra}(\theta)$ is the intra-video ranking loss that encourages queries to be closer to the corresponding video moment than all other possible moments from the same video, $L_i^{inter}(\theta)$ is the inter-video ranking loss that encourages queries to be closer to corresponding video moments than moments with the same endpoints outside the video, $L_R(x,y)=max(0,x-y+b)$ is the ranking loss, $D$ is the square distance metric, $\theta$ refers to the learnable parameters of the model, and $\lambda$ is the weighting parameter. Building on top of MCN, several researchers promoted this branch from different angles, such as visual and query modeling. In the following, we would detail the corresponding studies in sequence.

\textbf{Visual Modeling.} 
 \cite{Lisa2017Localizing} considers the entire video as the visual context, which may induce inappropriate information in the context feature, influencing the localization performance. Inspired by this, Hendricks et al.~\cite{Lisa2018Localizing} proposed the Moment Localization with Latent Context (MLLC) network\footnote{\url{https://people.eecs.berkeley.edu/}.}, which models visual context as a latent variable. Therefore, MLLC could attend to different video contexts conditioned on the specific query-video pair, overcoming the limitation of query-independent contextual reasoning. Besides, \cite{Lisa2018Localizing} builds the temporal reasoning in video and language dataset (TEMPO) based on the DiDeMo dataset~\cite{Lisa2017Localizing}, which comprises two parts: TEMPO-TL (Template Language) and TEMPO-HL (Human Language). The former is constructed by the original DiDeMo sentences with language templates, while the latter is built with real videos and newly collected user-provided temporal annotations.

\textbf{Query Modeling.} A common limitation of previous approaches is that they utilize a single textual embedding to represent the query, which may not be precise. To fully exploit temporal dependencies between events within the query, Zhang et al.~\cite{Songyang2019Exploiting} proposed a Temporal Compositional Modular Network (TCMN)\footnote{\url{https://github.com/Sy-Zhang/TCMN-Release}.}, which leverages a tree attention network to parse the given query into three components concerning the main event, context event, and temporal signal. Besides, a temporal relationship module is designed to measure the similarity between the phrase embedding for the temporal signal and the location feature, and a temporal localization module is designed to measure the similarity between the event-related phrase embedding and the visual feature.
The overall matching score between each moment candidate and the given query is calculated by the late fusion strategy.

\subsubsection{\textbf{Multi-scale sliding windows}}
The work of Gao et al.~\cite{Jiyang2017TALL} is the pioneer in this branch. They proposed a Cross-modal Temporal Regression Localizer (CTRL) to tackle the video moment localization task\footnote{\url{https://github.com/jiyanggao/TALL}.}. To collect moment candidates for training, CTRL adopts multi-scale temporal sliding windows. In the framework of CTRL, the pre-trained C3D model~\cite{Tran_2015_ICCV} is first utilized to extract visual features for moment candidates. Meanwhile, a Long Short-term Memory (LSTM)~\cite{Hu_2016_CVPR} network or Skip-thought~\cite{NIPS2015_5950} is adopted to extract query representations. Afterwards, a cross-modal processing module is designed to jointly model the query and visual features, which calculates element-wise addition, multiplication, and direct concatenation. Finally, the multi-layer perception network (MLP) is designed for visual semantic alignment and moment location regression. To jointly train for visual-semantic alignment and moment location regression, CTRL utilizes a multi-task loss as follows,
\begin{equation}
\begin{cases}
     &L=L_{aln}+\alpha L_{reg},\\
     &L_{aln}=\frac{1}{N}\sum_{i=0}^{N}[\alpha_c log(1+exp(-cs_{i,i}))+\sum_{j=0,j\neq i}^{N}\alpha_wlog(1+exp(cs_{i,j}))],\\
     &L_{reg}=\frac{1}{N}\sum_{i=0}^{N}[R(t_{s,i}^*-t_{s,i})+R(t_{e,i}^*-t_{e,i})],\\
\end{cases}   
\end{equation}
where $N$ is the batch size, $cs_{i,j}$ is the matching score between query $s_{j}$ and moment candidate $m_{i}$, $\alpha_c$ and $\alpha_w$ are the hyper parameters, $R(t)$ is the $L_1$ function, $t_{s,i}^*$ and $t_{e,i}^*$ is the ground truth offsets, as well as $t_{s,i}$ and $t_{e,i}$ is predicted offsets.
More importantly, for evaluation, CTRL adopts the TACoS dataset, and builds a new dataset on top of Charades by adding sentence temporal annotations, called Charades-STA. 
Hereafter, to further improve the localization performance from different aspects (e.g., visual, query, and interaction modeling), several methods are proposed in the past few years. We will elaborate them in the following paragraphs.

\textbf{Visual Modeling.}
Although CTRL considers visual context information, it ignores the complex interactions among contexts and fails to identify the importance of each moment, therefore some important cues are missing. Inspired by this, Liu et al.~\cite{Meng2018Attentive} extended the work of~\cite{Jiyang2017TALL} and developed an Attentive Cross-Modal Retrieval Network (ACRN)\footnote{\url{https://sigir2018.wixsite.com/acrn}.}. In particular, they designed a memory attention mechanism to emphasize the visual features mentioned in the query, obtaining the augmented moment representations. 

As visual feature extracted from the $Fc6$ layer of C3D may weaken or ignore the critical visual cues, a Visual-attention Action Localizer (VAL) is introduced in \cite{Xiaomeng2018VAL}, which extracts visual features from the $CONV5\_3$ layer of C3D. Therefore, it could capture more completed visual information. Similarly, Jiang et al.~\cite{Bin2019Cross} proposed a Spatial and Language-Temporal Attention (SLTA) method, which takes advantage of object-level local features to enhance the visual representations\footnote{\url{https://github.com/BonnieHuangxin/SLTA}.}. 
To be specific, SLTA extracts local features on each frame by Faster R-CNN~\cite{NIPS2015_5638} and introduces the spatial attention to selectively attend to the most relevant local features mentioned in the query. And then it utilizes LSTM to encode the local feature sequence, obtaining the local interaction feature. Finally, SLTA integrates the global motion features extracted by pre-trained C3D with local interaction features as the final visual representations. 
Recently, to identify the fine-grained differences among similar video moment candidates, Zeng et al.~\cite{mmrg} proposed a Multi-Modal Relational Graph (MMRG) framework\footnote{\url{https://cvpr-2021.wixsite.com/mmrg.}}. It develops a duel-channel relational graph to capture object relations and the phrase relations. Moreover, it considers two self-supervised pre-training tasks: attribute masking and context prediction, to enhance the visual representation and alleviate semantic gap across modalities.

\textbf{Query Modeling.}
CTRL simply treats queries holistically as one feature vector, which may obfuscate the keywords that have rich temporal and semantic cues. Inspired by this, Liu et al.~\cite{liumm2018} proposed a cRoss-modal
mOment Localization nEtwork (ROLE)\footnote{\url{https://acmmm18.wixsite.com/role}.}. 
It designs a language-temporal attention module, which adaptively reweighs each word’s features according to the textual query information and moment context information, therefore deriving useful query representations.

\textbf{Interaction Modeling.}
Existing scheme CTRL adopts the straightforward multi-modal fusion method, i.e., fusing element-wise addition, multiplication, and concatenation, lacking in-depth analysis. 
To overcome this drawback, Wu et al.~\cite{Aming2018Multi} put forward a new multi-modal fusion approach, i.e., Multi-modal Circulant Fusion (MCF)\footnote{\url{https://github.com/AmingWu/Multi-modal-Circulant-Fusion}.}. Particularly, after reshaping feature vectors into circulant matrices, it defines two types of interaction operations between vectors and matrices. 
The first one is the matrix multiplication between circulant matrix and projection vector, while the second is element-wise product between projection vector and each row vector of circulant matrix. More importantly, the proposed MCF can be integrated into the existing video moment localization models, to further improve the localization accuracy. 
To adequately exploit rich semantic cues about activities in videos and queries, Ge et al.~\cite{Runzhou2019MAC} proposed an Activity Concepts based Localizer (ACL)\footnote{\url{https://github.com/runzhouge/MAC}.}, which is the first work to mine the activity concepts from both the videos and the sentence queries to facilitate the localization. 
To explore the interactions between two modalities, ACL separately conducts the multi-modal processing to the pair of activity concepts and the pair of visual features and sentence embeddings.   
Besides, ACL designs an actionness score generator to calculate the likelihood of the moment candidate containing meaningful activities. Eventually, the alignment score multiplied by the actionness score is set as the final score to predict the alignment confidence between each candidate and query. 

Different from aforementioned two methods that focus on inter-modal interaction modeling,  Zhang et al.~\cite{migcn} proposed a Multi-modal Interaction Graph Convolutional Network (MIGCN)\footnote{\url{https://github.com/zmzhang2000/MIGCN/.}}, which simultaneously explores the complex intra-modal relations and
inter-modal interactions residing in the video and sentence query. Particularly, it introduces the
multi-modal interaction graph, where two types of nodes are considered (i.e., clip nodes and word nodes) and edges compile both intra-modal relations and inter-modal interactions. Therefore, the graph convolution could refine node representations by jointly considering intra- and inter-modal interaction information.

\subsubsection{\textbf{Moment generation networks}}
Instead of using hand-crafted heuristics or multi-scale sliding windows to generate moment candidates, Xu et al.~\cite{Huijuan2019Multilevel} advanced a Query-guided Segment Proposal Network (QSPN) to generate moment candidates\footnote{\url{https://github.com/VisionLearningGroup/Text-to-Clip_Retrieval}.}.  
Concretely, QSPN generalizes the SPN from R-C3D
model~\cite{Xu_2017_ICCV} by introducing query representations as the guidance information to generate moment candidates. And an early fusion retrieval model, instantiated as a two-layer LSTM, is introduced  to find the moment that best matches the given query. 
Besides, a captioning loss is considered to enforce the LSTM to re-generate the query sentence, achieving improved retrieval performance. The combination of retrieval loss and captioning loss is defined as follows, 
\begin{equation}
    \begin{cases}
    &L=L_{ret}+L_{cap},\\
    &L_{ret} = \sum_{j}max\{0, \eta+\sigma(s_j,m_j^{'})-\sigma(s_j,m_j)\},\\
    &L_{cap}=-\frac{1}{KT}\sum_{k=1}^{K}\sum_{t=1}^{T_k}logP(w_t^k|f(m_k),h_{t-1}^{(2)},w_1^k,\ldots,w_{t-1}^k),
    \end{cases}
\end{equation}
where $\sigma(s_j,m_j)$ is the matching score predicted by the LSTM, $m_j^{'}$ is the negative moment either comes from the same video or from a different video, $K$ is the number of queries, $T_k$ is the number of words in the $k$-th query, $f(m_k)$ represents the visual feature of the moment aligned with the $k$-th query. Similarly, Chen et al.~\cite{Shaoxiang2019Semantic} proposed a Semantic Activity Proposal (SAP) framework, which also integrates the semantic information in queries into the moment candidate generation process. To be specific, it first trains a visual concept detection CNN with paired query-clip training data. 
Subsequently, the visual concepts extracted from the query and video frames are utilized to calculate the visual-semantic correlation score for every frame. By grouping frames with high visual-semantic correlation scores, moment candidates could be generated. 
Different from the above two methods, Xiao et al.~\cite{BPNet} proposed a Boundary Proposal Network (BPNet), which utilizes VSLNet~\cite{Hao2020Span} to generate several high-quality moment proposals, avoiding redundant candidates.

\textbf{Summarization.} 
In what follows, we summarize the two-stage supervised video moment retrieval methods.
\begin{itemize}
    \item \textit{Hand-crafted heuristics}: In this branch, the pioneer work MCN focuses on learning the shared embedding space for the moment representation and the query representation. To further improve the performance of MCN,  MLLC enforces the model to attend to different video contexts, while TCMN exploits temporal dependencies between events in the  query.  

\item  \textit{Multi-scale sliding windows}: 
To enhance the performance of the pioneer work CTRL in this branch, ACRN, VAL, SLTA, and MMRG focus on the visual  modeling. Specifically, ACRN emphasizes the  visual features mentioned in the query, VAL captures more completed visual information from the feature map of C3D, SLTA takes advantage of object-level local features, and MMRG considers self-supervised pre-training tasks.   
Differently, ROLE focuses on query modeling and designs a language-temporal attention module to derive useful query representations. Moreover, MCF, ACL, and MIGCN pay attention to the interaction modeling. Particularly, MCF and ACL are devoted to modeling inter-modal interactions, while MIGCN jointly considers the inter-modal and intra-modal interaction modeling. 

\item \textit{Moment  generation  networks}: To generate moment candidates, QSPN generalizes the SPN  by introducing query representations as the guidance information, while SAP generates moment candidates by  grouping  frames with high  visual-semantic  correlation  scores. Differently,  BPNet directly utilizes the VSLNet~\cite{Hao2020Span} to generate moment proposals.
\end{itemize}

\begin{figure}
    \centering
    \includegraphics[width=0.5\textwidth]{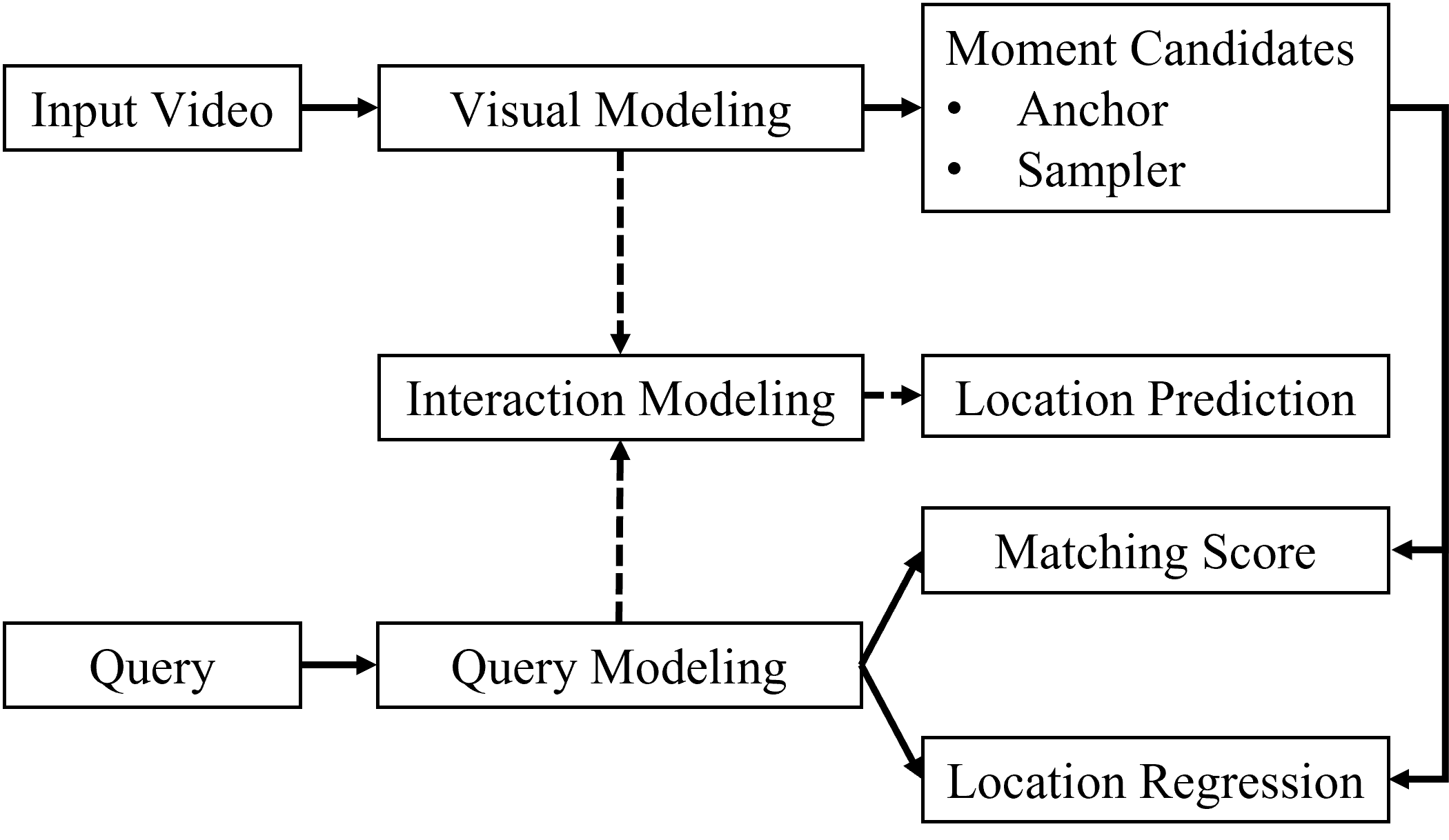}
    \vspace{-2ex}
    \caption{A block diagram of one-stage video moment localization.}
    \label{fig:onestage}
\end{figure}

\subsection{One-stage Methods}
Although two-stage approaches have achieved promising performance, they are suffering from inferior efficiency. For instance, to handle the diverse temporal scales and locations of the moment candidates, exhaustive matching between a large number of overlapping moments and the query is required, which is very computationally expensive. Therefore, developing one-stage methods that localize the golden moment (i.e., the moment that best matches the given query) in one single pass is critical. 

As summarized in Table~\ref{tab1}, according to the processing of moment candidates, existing one-stage or single-pass video moment localization methods can be grouped into three categories: anchor-based, sampler-based, and proposal-free ones. A block diagram of typical one-stage methods is shown in Fig.~\ref{fig:onestage}. In what follows, we review them in sequence.

\subsubsection{\textbf{Anchor-based models}} 
Chen et al.~\cite{Jingyuan2018Temporally} proposed the first dynamic single-stream end-to-end Temporal GroundNet (TGN) model for video moment localization\footnote{\url{https://github.com/JaywongWang/TGN}.}. 
It sets each time step as an anchor and generates $K$ moment candidates that end at the current time step. Moreover, it utilizes the hidden states generated by the interaction module to yield confidence scores for moments ending at the corresponding time step.
The objective of TGN is defined as,
\begin{equation}
    L=\sum_{(V, S)}\sum_{t=1}^{T}\{-\sum_{k=1}^{K}w_0^ky_t^klogc_t^k+w_1^k(1-y_t^k)log(1-c_t^k)\},
\end{equation}
where $(V,S)$ denotes a training pair from the training set, $y_t^k$ is interpreted as whether the $k$-th moment candidate at time step $t$ corresponds to the given query, $c_t^k$ denotes the prediction result, $w_0^k$ and $w_1^k$ are calculated according to the frequencies of positive and negative samples in the training set with length $k\delta$.

According to the primary contributions of existing anchor-based one-stage approaches, they can be further classified into two categories, i.e.,  focusing on localization module building and interaction modeling.

\textbf{Localization Module.}
To yield more precise localization, Wang et al.~\cite{Jingwen2020The} proposed an end-to-end Contextual Boundary-aware Prediction (CBP) model\footnote{\url{https://github.com/JaywongWang/CBP}.}, which jointly predicts temporal anchors and boundaries at each time step. Its localization module consists of an anchor sub-module as~\cite{Jingyuan2018Temporally} did and a boundary sub-module that is utilized to decide whether the current step is a semantic boundary corresponding to the start/end time of the moment. 

\textbf{Interaction Modeling.}
As simply combining the video and query information is not expressive enough to explore the interaction information, Li et al.~\cite{Cheng2019Bidirectional} proposed a Bidirectional Single-Stream Temporal Localization (BSSTL) model.
It introduces an attentive cross-modal fusion model, consisting of dynamic attention weighted query, joint gating, and fusion, to capture the interaction between the queries and videos. 
Liu et al.~\cite{liu-etal-2020-reasoning} designed a deep Rectification-Modulation Network (RMN), which adopts a multi-step reasoning framework to gradually capture the higher-order multi-modal interaction. To be specific, it utilizes multi-step rectification-modulation layers to progressively refine video and query interactions, improving localization accuracy. 
Different from previous approaches that merely focus on exploiting unidirectional interaction information from the video to query,  
 Qu et al.~\cite{Xiaoye2020Fine} proposed a Fine-grained Iterative Attention Network (FIAN) to capture bilateral query-video interaction information.
 Specifically, it leverages a symmetrical iterative attention module to generate video-aware query and query-aware video representations, where two attention branches share the same architecture but opposite input.
To model multi-level vision-language cross-modal relation and long-range context, Ning et al.~\cite{iin} proposed an Interaction-Integrated Network (I2N). To be specific, it introduces an  interaction-integrated cell, which integrates both cross-modal and contextual interactions between vision and language. By stacking a few interaction-integrated cells in a chain, the network could 
explore more detailed vision-language relations on multiple granularities, therefore achieving more accurate semantics understanding and more accurate localization results.

The aforementioned single-pass methods mainly focus on exploring the cross-modal relations between the video and query, ignoring the importance of jointly investigating the cross- and self-modal relations. In light of this, 
Zhang et al.~\cite{Zhu2019Cross} introduced a Cross-Modal Interaction Network (CMIN)\footnote{\url{https://github.com/ikuinen/CMIN}.}, which jointly considers multiple crucial factors. Specifically, 
the syntactic graph convolution network 
is adopted to enhance the query understanding. And the multi-head self-attention module is built to capture long-range semantic dependencies from the video context. Moreover, a multi-stage cross-modal interaction
module is presented to exploit the potential relations of video and query contents. Recently, inspired by the fact that captioning can improve the performance
of image-based grounding of textual phrases~\cite{rohrbach2016grounding}, Zhang et al. further improved the model CMIN by integrating a query reconstruction module~\cite{cmin_ext}, dubbed as CMIN-R. 
Likewise, Liu et al.~\cite{Daizong2020Jointly} presented a Cross- and Self-Modal Graph Attention Network (CSMGAN)\footnote{\url{https://github.com/liudaizong/CSMGAN}.}. 
It introduces a cross-modal relation graph to highlight relevant instances across the video and query. Meanwhile, a self-modal relation graph is utilized to model the pairwise relation inside each modality. In the same year, Chen et al.~\cite{Shaoxiang2020Learning} proposed a Pairwise Modality Interaction module (dubbed PMI-LOC), which feeds multimodal features to a channel-gated modality interaction module, to exploit both intra- and inter-modality information.

\subsubsection{\textbf{Sampler-based models}} 
Different from anchor-based approaches that generate moment candidates via multi-scale windows at each time step, sampler-based ones output moment candidates via different sampler networks, such as temporal convolution, segment-tree, and enumeration.

\textbf{Temporal Convolution.}
Yuan et al.~\cite{Yitian2019Semantic} proposed a Semantic Conditioned Dynamic Modulation (SCDM) mechanism based approach\footnote{\url{https://github.com/yytzsy/SCDM}.}, which leverages query semantic information to modulate the temporal convolution processes in a hierarchical temporal convolutional network. Particularly, 
similar to \cite{lin17}\cite{Liu_2016} for object/action detection, the position prediction module is introduced to output location offsets and overlap scores of moment candidates based on the modulated features. 
Similarly, Zhang et al.~\cite{Da2019MAN} proposed Moment Alignment Network (MAN)\footnote{\url{https://github.com/dazhang-cv/MAN}.} to deal with semantic misalignment. In particular, it treats the encoded word features as efficient dynamic filters to convolve with input visual representations, and then a hierarchical convolutional network is applied to directly produce multi-scale moment candidates. Different from MAN, Hu et al.~\cite{clear} presented an
end-to-end Coarse-to-fine cross-modaL sEmantic Alignment netwoRk (CLEAR)\footnote{\url{https://github.com/Huyp777/CSUN.}}, which utilizes bi-directional temporal convolution network followed by multi-scale temporal pooling to generate moment candidates. And to explore cross-modal semantic correlation and improve the localization efficiency, it respectively advances a multi-granularity interaction module and a semantic pruning
strategy.

Different from that existing methods mainly leverage vanilla soft attention to perform the
alignment in a single-step process, Liu et al.~\cite{ia-net} presented an Iterative Alignment Network (IA-Net), which captures complicated relations between inter- and intra-modality 
through multi-step reasoning. To be specific, it proposes an improved co-attention mechanism that utilizes learnable paddings to address nowhere-to-attend problem with deep latent clues, and a calibration module to refine the alignment knowledge of inter- and intra-modal relations.

To boost the efficiency of moment localization, Hu et al.~\cite{cmhn} proposed an
end-to-end Cross-Modal Hashing Network (CMHN)\footnote{\url{https://github.com/Huyp777/CMHN.}}. It designs a cross-model hashing module to project cross-modal heterogeneous representations into a shared isomorphic Hamming space for compact hash code learning. Therefore, with the well-trained model at hand, the hash codes of any upcoming videos could be obtained offline and independently, improving the localization efficiency and scalability.

\textbf{Segment-tree.}
Unlike the aforementioned sampler-based approaches, Zhao et al.~\cite{cpn} formulated the video moment localization task as a multi-step decision problem and proposed a Cascaded Prediction Network (CPN). To be specific, it adopts a segment
tree-based structure to generate moment candidates in different temporal scales and refine the representation of them in a message-passing way via graph neural network.

\textbf{Proposal Generation Network.}
To well exploit moment-level interaction and speed up the localization efficiency, Liu et al.~\cite{APGN} proposed an Adaptive Proposal Generation Network (APGN). It first leverages a binary classification module to predict foreground frames, and then it utilizes the boundary regression module to generate proposals on each foreground frame. In this way, the redundant proposals are decreased. Meanwhile, Xiao et al.~\cite{LP-Net} proposed a Learnable Proposal Network (LP-Net) for video moment localization\footnote{\url{https://github.com/xiaoneil/LPNet/.}}. In the LP-Net, proposal boxes are represented by 2-d parameters ranging from 0 to 1, denoting normalized center coordinates
and lengths, which are randomly initialized. And these learnable proposal boxes are updated by dynamic adjustor during training. 

\textbf{Enumeration.}
Given that the input video has temporal length $n$, the Temporal Modular Network (TMN) proposed in~\cite{Bingbin2018Temporal} will regress $n$ correspondence scores for each temporal segment, and then combine the scores of consecutive segments to produce $\frac{n(n+1)}{2}$ scores for all possible moment candidates. Finally, the sub-video with the maximum score is predicted as the golden moment. 

To model temporal relations between video moments, Zhang et al.~\cite{Songyang2020Learning} presented a 2D temporal map, where one dimension indicates the start time of a moment and the other indicates the end time. Based on the 2D map, \cite{Songyang2020Learning} introduces a Temporal Adjacent Network (2D-TAN)\footnote{\url{https://github.com/microsoft/2D-TAN}.} to generate the 2D score map, i.e., the matching scores of moment candidates on the 2D temporal map with the given query. Recently, Zhang et al.~\cite{MS-2D-TAN} extended the 2D-TAN to a
multi-scale version and proposed a Multi-Scale Temporal Adjacency Network (MS-2D-TAN), which models the temporal context between video moments by a set of predefined two-dimensional maps under different temporal scales\footnote{\url{https://github.com/microsoft/2D-TAN.}}.

Different from aforementioned methods that use alignment information to find out the best-matching candidate, Wang et al.~\cite{DIN} proposed a unified Dual Path Interaction Network (DPIN). It jointly considers the alignment and discrimination information to make the prediction. Particularly, the frame-level representation path extracts the discriminative boundary information from the fused features, while the candidate-level path arranges the moment candidate features in a 2D temporal map as \cite{Songyang2020Learning} did and extract the alignment information. Finally, DPIN fuses the two kinds of representations to make prediction. Likewise, Wang et al.~\cite{smin} developed a Structured Multi-level Interaction Network (SMIN) by jointly considering multiple levels of visual-textual interaction and moment structured interaction. 
\cite{matn} presents the Multi-stage Aggregated Transformer Network (MATN) to enhance the cross-modal alignment and localization accuracy. Particularly, it designs a new visual-language transformer backbone by using different parameters to process different modality contents, and introduces a multi-stage aggregation module to calculate the moment representation via considering three stage-specific representations. 
Wu et al.~\cite{sv-vmr} proposed a video moment retrieval model, named SV-VMR, which jointly models the fine-grained and comprehensive
relations by using both semantic and visual structures. 
Unlike previous work, Gao et al.~\cite{ranet} formulated the video moment localization task as
the video reading comprehension and presented a Relation-aware Network (RaNet)\footnote{\url{https://github.com/Huntersxsx/RaNet.}}. To distinguish similar moment candidates in the visual modality, it introduces a coarse-and-fine cross-modal interaction module to simultaneously capture the sentence-moment and token-moment level interaction, obtaining sentence-aware and token-aware moment representations. Meanwhile, it leverages graph convolutional network to capture moment-moment relations, which further strengthens the discriminative of moment representations. 

To against the temporal location biases of video moment localization, Yang et al.~\cite{dcm} advanced the Deconfounded Cross-modal Matching (DCM) method\footnote{\url{https://github.com/Xun-Yang/Causal_Video_Moment_Retrieval.}}, which considers the moment temporal location as a hidden confounding variable. To remove the confounding effects of moment location, it disentangles the moment representation to infer the core feature
of visual content, and then applies causal intervention on the multi-modal input based on backdoor adjustment. 
To well balance the localization accuracy and speed, Gao et al.~\cite{fvmr} proposed the Fast Video Moment Retrieval (FVMR) model, which replaces the complex cross-modal interaction module in existing methods with a cross-modal common space.

\subsubsection{\textbf{Proposal-free models}} 
Unlike the anchor- and sampler-based methods that depend on moment candidates, the proposal-free ones directly predict the start and end time of the target moment. This removes the need to retrieve and re-rank multiple moment candidates.

As the first work in this subbranch, Chen et al.~\cite{Jingyuan2019Localizing} proposed a localizing network (LNet), which works in an end-to-end fashion, to tackle the video moment localization task. It first matches the query and video sequence by cross-gated attended recurrent networks, to exploit their fine-grained interactions and generate a query-aware video representation. Afterwards, a self interactor is designed to perform cross-frame matching, which dynamically encodes and aggregates the matching evidences. Finally, a boundary model is introduced to locate the positions of video moments corresponding to the query by directly predicting the start and end points. To further improve the localization accuracy of LNet, the follow-up research of this branch is dedicated to exploring how to well exploit interactive information between the video and the given query.

To adequately explore the cross-modal interactions between the query and video, Yuan et al.~\cite{Yitian2019To} proposed an end-to-end Attention Based Location Regression (ABLR) model \footnote{\url{https://github.com/yytzsy/ABLR_code}.}. It leverages a multi-modal co-attention mechanism to learn both video and query attentions.
Moreover, a multi-layer attention-based location prediction network is proposed to regress the temporal coordinates for the target video moment. Similarly, Ghosh et al.~\cite{Soham2019ExCL} proposed an Extractive Clip Localization (ExCL) approach to predict the start and end frames of the target moment. 
Particularly, they compared three variants of span predictor, i.e., MLP, Tied-LSTM, and Conditioned-LSTM, to predict the start and end probabilities for each frame. Liu et al.~\cite{SSMN} designed a Single-shot Semantic Matching Network (SSMN), which properly explores the cross-modal relationship and the temporal information via an enhanced cross-modal attention module. For in-depth relationship modeling between semantic phrases and video segments, Mun~\cite{Jonghwan2020Local} proposed a Sequential Query Attention Network (SQAN) based method by performing local-global video-query interactions\footnote{\url{https://github.com/JonghwanMun/LGI4temporalgrounding}.}.  
Tang et al.~\cite{acrm} proposed an Attentive Cross-modal Relevance Matching (ACRM) model\footnote{\url{https://github.com/tanghaoyu258/ACRM-for-moment-retrieval.}}, which uses element-wise
multiplication and subtraction functions to model the interaction between the frame and frame-specific query features. Liang et al.~\cite{liang2021local} proposed a Multi-branches Interaction Model (MIM), which re-weights the video features according to relevance of query and video in multiple sub-spaces. By treating the video as a text passage and the target moment as the answer span, Zhang et al.~\cite{Hao2020Span} proposed a video span localizing network (VSLNet) on top of the standard span-based QA framework\footnote{\url{https://github.com/IsaacChanghau/VSLNet}.}. In particular,  a context-query attention (CQA) \cite{xiong2016dynamic} is designed to capture the cross-modal interactions between visual and textural features. Recently, to address the performance degradation on long videos, Zhang et al.~\cite{VSLNet-L} further extended VSLNet to VSLNet-L by introducing a multi-scale split-and-concatenation strategy. It first partition long video into clips of different lengths, and then locate the target moment in the clips that are more likely to contain it. Unlike the above methods, Opazo et al.~\cite{Cristian2020Proposal} introduced a Proposal-free Temporal Moment Localization model using Guided Attention (PFGA)\footnote{\url{https://github.com/crodriguezo/TMLGA}.}, which utilizes an attention-based dynamic filter to transfer query information to the video. Moreover, a new loss is designed to enforce the model to focus on the most relevant part of the video. 
Since humans have difficulty agreeing on the start and end
time of an action inside a video~\cite{Alwassel_2018_ECCV}\cite{Sigurdsson_2017_ICCV}, 
PFGA uses soft-labels~\cite{Szegedy_2016_CVPR} to model the uncertainty associated to the labels. The final loss for training PFGA method is defined as,
\begin{equation}
    \begin{cases}
    &L=L_{KL}+L_{att},\\
    &L_{KL}=D_{KL}(\hat{\tau}^s||\tau^s)+D_{KL}(\hat{\tau}^e||\tau^e),\\
    &L_{att}=-\sum_{i=1}^{n}(1-\delta_{\tau^s\leq i \leq \tau^e}log(1-a_i),
    \end{cases}
\end{equation}
where $L_{KL}$ loss aims to minimize the Kullback-Leibler divergence between the predicted and ground truth probability distributions, $L_{att}$ loss is designed to encourage the model to attend relevant features, $\delta$ is the Kronecker delta, $n$ denotes the length of visual sequence, and $a_i$ is the attention of the corresponding location.

To further explore fine-grained interaction information within the video and query sentence, Chen et al.~\cite{HTV} advanced a Hierarchical Visual-Textual Graph (HVTG) model for video moment localization\footnote{\url{https://github.com/forwchen/HVTG}.}.
To be specific, it first builds an object-sentence subgraph to obtain sentence-aware object features for each frame. And then it feeds sentence-aware object features into an object-object subgraph, to capture object-object interactions inside each frame. Afterwards, a sentence-guided attention followed by a Bi-LSTM is introduced to 
establish temporal relations among frames, outputting final visual representations. To explore multi-level query semantics (both word- and phrase-level) as well as model multi-level interactions between two modalities, Tang et al.~\cite{mqei} proposed a  Multi-level Query Exploration and Interaction (MQEI) model. To be specific, it leverages a stack of syntactic GCNs to model the syntactic dependencies of the query, obtaining the word-level query features. And a sequential attention module is adopted to learn phrase-level query representations. With such multi-level query representations, the context-query attention is adopted for the word-level and phrase-level cross-modal feature fusion.

Different from most proposal-free methods that are devoted to multi-modal modeling (i.e., fusion or interaction), Li et al.~\cite{cp-net} focused on exploiting the fine-grained temporal clues in videos and presented a Contextual Pyramid Network (CP-Net) to mine rich temporal contexts through
fine-grained hierarchical correlation at different 2D temporal scales. Similarly, a language-conditioned message-passing
algorithm is proposed in \cite{dori} to well learn the video feature embedding, which could capture the relationships between humans, objects and activities in the video. Inspired by the successful application of biaffine mechanism in dependency parsing, Liu et al.~\cite{cbln} applied the biaffine mechanism to video moment localization and presented the Context-aware Biaffine Localizing Network (CBLN)\footnote{\url{https://github.com/liudaizong/CBLN.}}. It leverages a multi-context biaffine localization module
to aggregate both multi-scale local and global contexts for each frame representation, and then scores all possible pairs of start and end frames for moment localization. 

To alleviate the imbalance problem between positive and negative samples to some extent, Lu et al.~\cite{Chujie2019DEBUG} proposed a DEnse Bottom-Up Grounding (DEBUG) framework by regarding all frames in the ground truth moment as positive samples. For each positive frame, DEBUG has a classification subnet to predict its relatedness with the query, and a boundary regression subnet to regress the unique distances from its location to bi-directional ground truth boundaries. 
This helps to avoid falling into the local optimum caused by independent predictions since each pair of boundary predictions is based on the same frame feature. Note that DEBUG can be seamlessly incorporated into any backbone to boost performance. 

Through utilizing the distances between the frame within the ground truth and the start (end) frame as dense supervisions, Zeng et al.~\cite{Runhao2020Dense} designed a dense regression network (DRN)\footnote{\url{https://github.com/Alvin-Zeng/DRN}.}. To be specific, DRN first forwards the video frames and the
query to the video-query interaction module for extracting the multi-scale feature maps. Afterwards, each feature map is processed by the grounding module to predict a temporal bounding box, a semantic matching score, and an IoU score at each temporal location for ranking. Finally, combining the matching score and the IoU score, DRN could find the best grounding result.  

\begin{figure}
    \centering
    \includegraphics[width=0.75\textwidth]{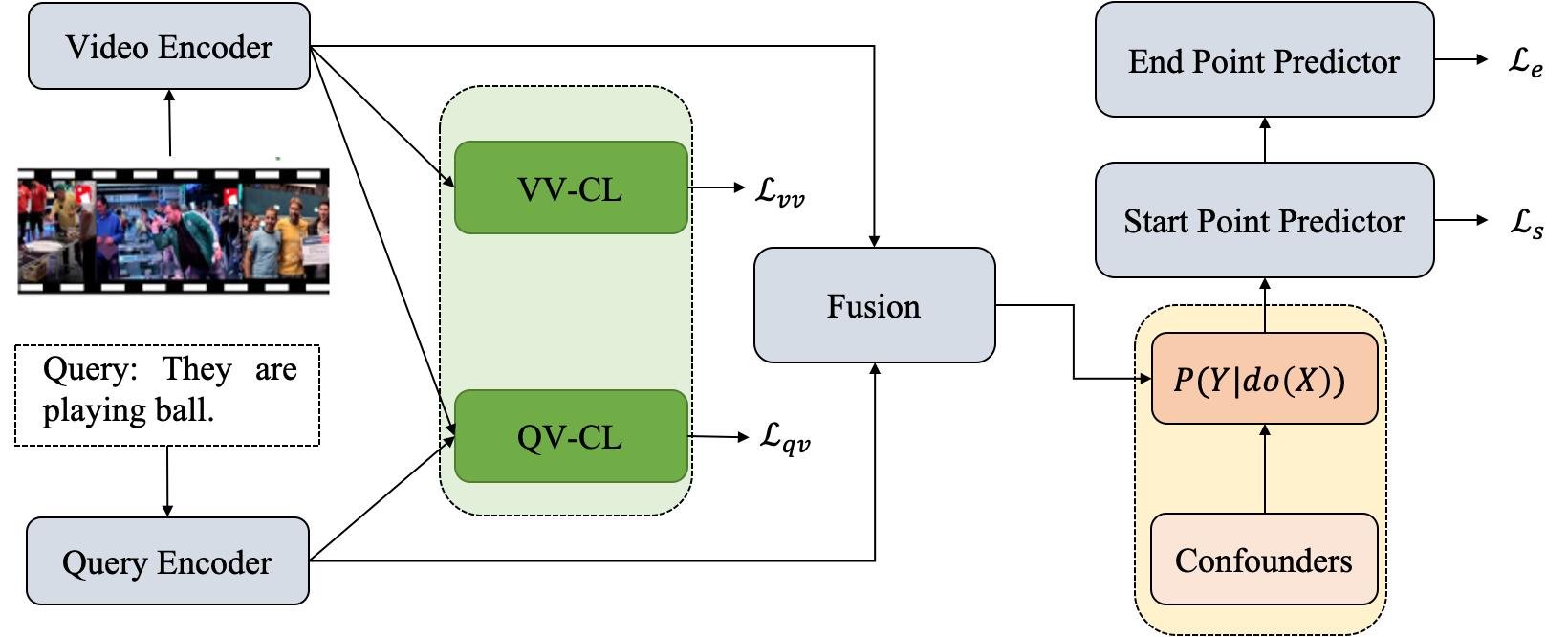}
    \vspace{-2ex}
    \caption{The architecture of IVG. VV-CL and
QV-CL respectively refer to two contrastive modules with losses expressed
as $L_{vv}$ and $L_{qv}$. $L_s$ and $L_e$ denote the cross-entropy
losses for predicting the boundary of the target span, figure from \cite{IVG}.}
    \label{fig:ivg}
    \vspace{-2ex}
\end{figure}

Despite the enormous success of the aforementioned models, they may suffer from the spurious correlations between textual and visual features due to
the selection bias of the dataset. To address this problem, ~\cite{IVG} first presents the Interventional Video Grounding (IVG) by introducing the causal interventions, as shown in Fig.~\ref{fig:ivg}. Moreover, it introduces a dual contrastive learning to learn more informative visual and textual representations. To be specific, the loss $L_{vv}$ is expressed as follows,
\begin{equation}
    L_{vv}=-I_{\theta}^{vv(s)}-I_{\theta}^{vv(e)},
\end{equation}
where $I_{\theta}^{vv(s)}$ and $I_{\theta}^{vv(e)}$ respectively denote the mutual information between the start and end boundaries of the video and the other clips. And the contrastive loss $L_{vq}$ is defined as follow,
\begin{equation}
    L_{vq}=-I_{\theta}^{vq}=-E_{V_a^{'}}[sp(C(q,V^{'}))]+E_{V_b^{'}}[sp(C(q,V^{'}))],
\end{equation}
where $V_a^{'}$ denote the features that reside within a target moment, $V_b^{'}$ denote the features are located outside of the target moment, $C$ refers to the mutual information discriminator, and $sp(z)=log(1+e^z)$.

Different from most existing methods that only consider RGB images as visual features, Chen et al.~\cite{drft} proposed a multi-modal learning framework for video temporal grounding
using (D)epth, (R)GB, and optical (F)low with the (T)ext as the query (DRFT). To model the interactions between modalities, this paper develops a dynamic fusion mechanism across modalities
via co-attentional transformer. Moreover, to enhance intra-modal feature representations, it leverages self-supervised contrastive learning across videos for each modality.

\textbf{Summarization.} Below, we summarize the one-stage video moment localization methods.
\begin{itemize}
    \item \textit{Anchor-based models:} In this branch, on top of TGN, CBP designs a boundary prediction module to predict temporal boundaries at each time step. To capture the interaction information, BSSTL, RMN, FIAN, and I2N focus on exploring the inter-modal interactions, while CMIN, CSMGAN, PMI-LOC and CMIN-R focus on exploring both intra- and inter-modal interactions.
    \item \textit{Sampler-based models:} In this branch, SCDM, MAN, CLEAR, IA-NEt, and CMHN adopt the temporal convolution network to generate moment candidates. Specifically, SCDM designs semantic modulated temporal convolution, MAN utilizes hierarchical convolution, and CLEAR leverages bi-directional temporal convolution. Although IA-Net and CMHN also utilizes temporal convolution to generate moment candidates, they respectively are devoted to tackling the interaction modeling and location efficiency issue. 
    CPN innovatively integrates the segment-tree into the video moment localization for moment generation. Differently, APGN and Lp-Net introduce different proposal generation networks for moment candidates generation. 
    TMN, 2D-TAN, MS-2D-TAN, DPIN, MATN, SMIN, SV-VMR, RaNet, DCM, and FVMR enumerate all possible moment candidates. More specifically, 2D-TAN and MS-2D-TAN respectively utilize the temporal  adjacent  network and multi-scale temporal  adjacent  network to gradually perceive more context of adjacent moment  candidates. DPIN, MATN, SMIN, SV-VMR, and RaNet focus on developing different mechanisms to  capturing interaction information.  Differently, DCM is devoted to addressing the location bias issue, while FVMR aims to well balance the location accuracy and efficiency. 
    \item \textit{Proposal-free models:} In this branch, 
    ABLR, ExCL, SQAN, PFGA, VSLNet, VSLNet-L, ACRM, MIM, and SSMN focus on exploring cross-modal interaction information between videos and queries. Thereinto, VSLNet and VSLNet-L tackle video moment localization by using the standard span-based QA framework. HVTG and MQEI are devoted to modeling both intra- and inter-modal interactions, while CBLN, CP-Net, and DORi merely pay attention to enhance the visual representation. DEBUG aims to tackle the imbalance problem between positive and negative samples, while DRN aims to promote the accuracy of localization by considering frame-level dense supervisions. IVG targets to address the spurious correlations between textual and visual features, while DRFT tackles the video moment localization by adopting multi-modal visual features.
\end{itemize}

\subsection{Reinforcement Learning}
As aforementioned, the one- and two-stage video moment localization studies inevitably suffer from inefficiency and unintelligent issues, supervised reinforcement learning based localization approaches are then proposed recently, as summarized in Table~\ref{tab1}.

\begin{figure}
    \centering
    \includegraphics[width=0.75\textwidth]{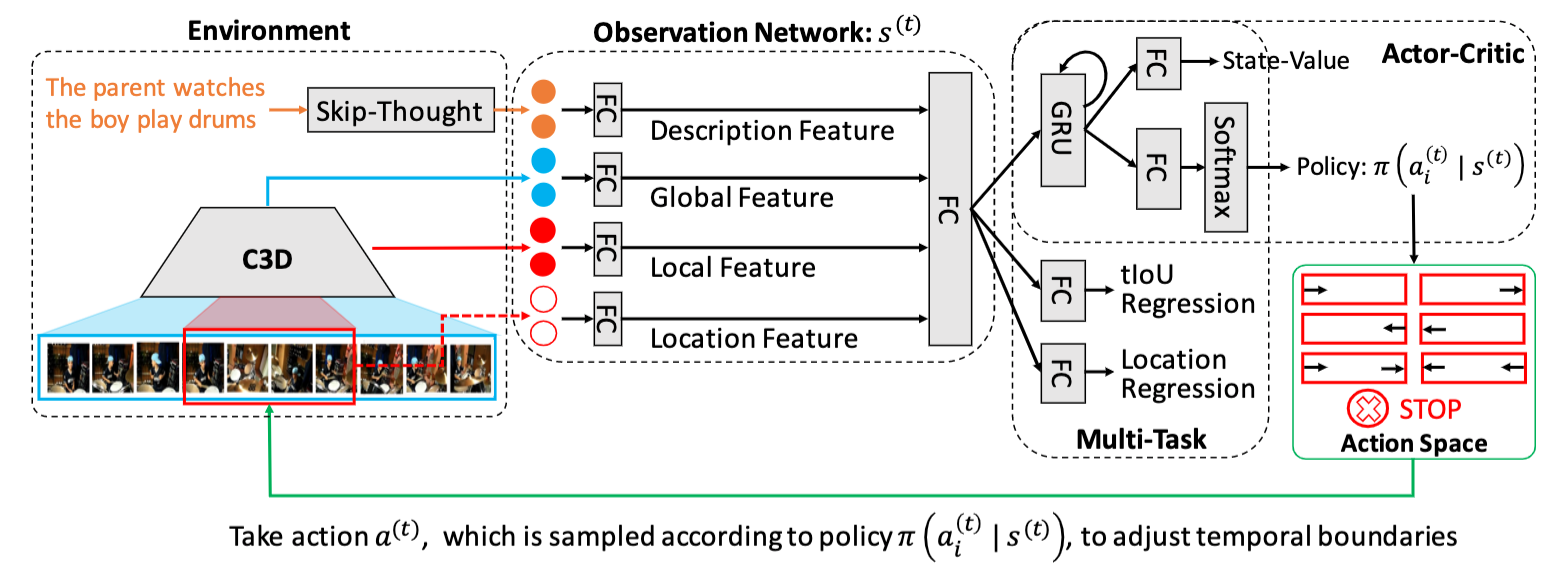}
    \vspace{-2ex}
    \caption{The overall pipeline of RWM model, figure from~\cite{Dongliang2019Read}.}
    \label{fig:rwm}
\end{figure}

He et al.~\cite{Dongliang2019Read} presented the first end-to-end reinforcement learning based framework, named RWM, which formulates the video moment localization as a problem of sequential decision making (as shown in Fig.~\ref{fig:rwm}). Concretely, 
its action space consists of 7 predefined actions\footnote{Moving start/end point ahead by $\delta$ (a predefined video length), moving start/end point backward by $\delta$, shifting both start and end point backward/forward by $\delta$, and a STOP action.}. At each time step, the observation network concatenates and fuses sentence embedding, global video feature, local video feature, and normalized temporal boundaries, generating the state vector. The obtained state vector is then fed into the actor-critic~\cite{sutton2018reinforcement} to learn the policy and state-value. To learn more representative state vectors, RWM combines reinforcement learning and supervised learning into a multi-task learning framework. 

Inspired by the coarse-to-fine decision-making paradigm of human, Wu et al.~\cite{Jie2020Tree} formulated a Tree-Structured Policy based Progressive Reinforcement Learning (TSP-PRL) framework\footnote{\url{https://github.com/WuJie1010/TSP-PRL}.}. 
Different from the previous work RWM~\cite{Dongliang2019Read}, it builds a hierarchical action space, containing all primitive actions in this task. 
Moreover, TSP-PRL designs a tree-structured policy to decompose complex action policies and obtains a reasonable primitive action via two stages selection. To optimize the tree-structured policy, the progressive reinforcement learning (PRL) is designed on the basis of~\cite{sutton2018reinforcement}.

To bridge the huge visual-semantic gap between videos and queries, Wang et al.~\cite{Weining2019Language} proposed a semantic matching reinforcement learning (SM-RL) model, which improves video representation by introducing visual semantic concepts.
Specifically, in the observation module, unlike previous methods merely leverage the global image information, the sentence query, as well as the location information to output the action and state values, SM-RL also integrates semantic concept features of videos. 
To jointly explore crucial clues hidden in the temporal and spatial information,  Cao et al.~\cite{strong} contributed a Spatio-Temporal ReinfOrcement learniNG (STRONG) framework for video moment localization\footnote{\url{https://github.com/yawenzeng/STRONG}.}. It first exploits a temporal-level
reinforcement learning to dynamically adjust the boundary of localized video moment. Thereafter, a spatial-level reinforcement learning is proposed to track the scene on consecutive image frames, therefore filtering out less relevant information. Concretely, the state is defines as the combination of the query sentence feature, the local video feature, and the spatial video feature. To simultaneously tackle the challenge of efficient search and query-video alignment, Hahn et al.~\cite{Meera2019Tripping} presented a Tripping through time Network (TripNet). Compared with \cite{Dongliang2019Read} and~\cite{Jie2020Tree}, the main contribution of TripNet is that it integrates the reinforcement learning (RL) and fine-grained video analysis. Concretely, in the state processing module, it designs a novel gated attention mechanism to model fine-grained textual and visual representations for video-text alignment. 
Besides, its reward function is defined as the difference of potentials between the previous state and current state. Therefore, the agent could find the target without looking at all frames of the video. 

Considering both limited moment selection and insufficient structural comprehension, Sun et al.~\cite{maban} proposed a Multi-Agent Boundary-Aware Network (MABAN)\footnote{\url{https://mic.tongji.edu.cn/e5/23/c9778a189731/page.htm.}}, which utilizes multi-agent reinforcement learning to obtain the two temporal boundary points for the target moment. At each step of temporal boundary adjustment, the start point agent and end point agent receive the state vector from the observation network and adjust the temporal boundaries in variable directions and scales, making moment selection more flexible and goal-oriented. To overcome the latter issue, a cross-modal interaction that considers semantic fusion in global and local phases is introduced to explore rich contextual information.

Different from pioneer methods that consider the video moment localization as either a ranking issue or a localization problem, Cao et al.~\cite{AVMR} proposed an Adversarial Video Moment Retrieval model (AVMR), which combines ranking and localization into a unified framework\footnote{\url{https://github.com/yawenzeng/AVMR}.}. 
To accelerate the convergence and increase the diversity of generated video moments, AVMR employs deep deterministic policy gradient (DDPG)~\cite{DDPG} to learn the policy. In addition, as utilizing indistinguishable values of IoU as the reward is unstable and difficult to convergence, AVMR introduces a discriminator to provide flexible reward, i.e., the Bayesian personalized ranking model.

\textbf{Summarization.}  
 In this category, TSP-PRL build a hierarchical action space and designs the progressive reinforcement learning to optimize its tree-structured policy. Different from TSP-PRL, TripNet focuses on enhancing visual representations, SM-RL introduces visual semantic concepts to calculate state values, while STRONG jointly considers the temporal-level and spatial-level  reinforcement  learning  to improve the localization performance. Differently, AVMR utilizes the Bayesian personalized ranking model to provide reward and MABAN is devoted to capturing cross-modal interactions. 

\begin{figure}[t]
	\centering
  \includegraphics[width=0.9\textwidth]{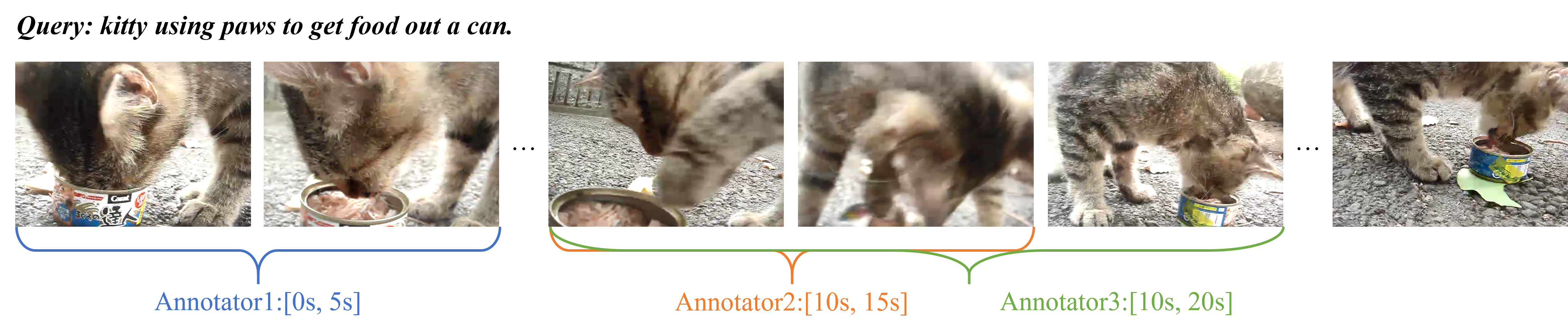}
  \vspace{-3ex}
\caption{An annotation example from the DiDeMo dataset. Each color refers to one annotation result of the given query.}\label{fig2}
\vspace{-3ex}
\end{figure}

\section{Weakly Supervised Video Moment Localization}
\textbf{Problem Statement.}
It is difficult for human to mark the start and end locations of a certain moment. As shown in Fig.~\ref{fig2}, three annotators give completely different annotation results for the same query and video. These inconsistent temporal annotations may introduce ambiguity in the training data. Moreover, acquiring dense annotations of query-temporal boundaries is often tedious. Intuitively, video-sentence pairs may be obtained with minimum human intervention as compared to temporal sentence annotations. Therefore, weakly supervised video moment localization task is introduced. Unlike supervised video moment localization that has access to the start and end time points of the queries, weakly supervised video moment localization only utilizes video-sentence pairs for training, namely it does not rely on temporal annotation information.

As summarized in Table~\ref{weak_m}, weakly supervised video moment localization studies can also be categorized into tree groups: two-stage, one-stage, and reinforcement learning based methods.

\begin{figure}
    \centering
    \includegraphics[width=0.75\textwidth]{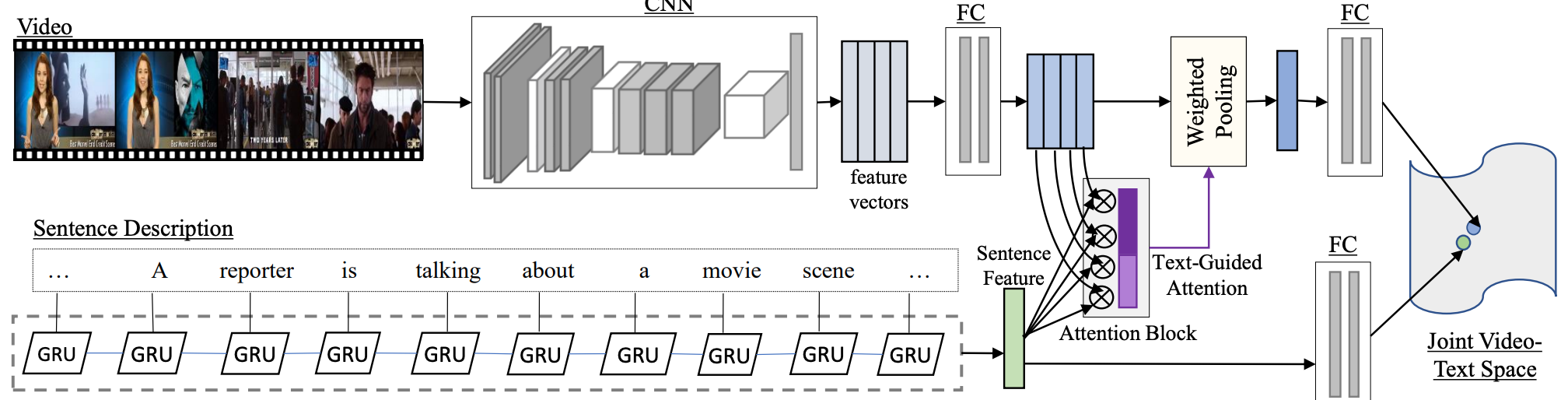}
    \vspace{-2ex}
    \caption{The overall framework of TGA, figure from~\cite{Niluthpol2019Weakly}}
    \label{fig:tga}
    \vspace{-3ex}
\end{figure}
\subsection{Two-stage Methods} 
To the best of our knowledge, the first weakly supervised video moment localization model is proposed in \cite{Niluthpol2019Weakly}\footnote{\url{https://github.com/niluthpol/weak_supervised_video_moment}.}. As shown in Fig.~\ref{fig:tga}, it aims to train a joint embedding network to project video and query features into the same joint
space. In particular, following~\cite{Jiyang2017TALL}, it first obtains the moment candidates via multi-scale sliding windows. 
After the feature extraction process, it introduces the Text-Guided Attention (TGA) module to obtain the query-wise video feature. Thereafter, two FC layers are utilized in \cite{Niluthpol2019Weakly} to project the query-specific video feature and paired query feature into the same joint space.

Inspired by the success of two-stream structure in weakly supervised object and action detection tasks \cite{bilen2016weakly}\cite{wang2017untrimmednets}, \cite{Mingfei2019Wslln} presents a Weakly Supervised Natural Language Localization Network (WSLLN), consisting of two branches: alignment branch and selection branch. The former branch is utilized to estimate the consistency score, while the latter one is adopted to calculate the selection score. 
Scores from both branches are then merged to produce the final score of each moment-query pair. Similarly, Ma et al.~\cite{Minuk2020VLANet} proposed the Video-Language Alignment Network (VLANet) to prune out spurious moment candidates. 
To be specific, a surrogate proposal selection module is designed to select the best-matched moment from each moment group based on the cosine similarity to the query embedding. 
Moreover, VLANet is trained  using contrastive loss that enforces semantically similar videos and queries to cluster in the joint embedding space. 

To improve the latent alignment between videos and natural language, Tan et al.~\cite{logan} designed a latent co-attention network (named LoGAN), which learns contextualized visual
semantic representations from fine-grained frame-by-word
interactions. Particularly, it builds a word-conditioned visual graph module  to learn contextualized visual semantic representations by integrating temporal contextual
information into the visual features.

\subsection{One-stage Methods}
Recently, Lin et al.~\cite{Zhijie2020Weakly} presented a Semantic Completion Network (SCN), which scores all the moments sampled at different scales in a single pass. 
Particularly, it introduces the semantic completion module to measure the semantic similarity between moments and the query, as well as  compute rewards. To train
the semantic completion module, SCN designs a reconstruction loss to force it to extract key information from the visual context.  
Different from previous works using paired sentences as supervision information, Wang et al.~\cite{vca} proposed the visual co-occurrence alignment (VCA) method, which 
targets to learn more discriminative and robust visual features by mining visual supervision information. Concretely, it utilizes the similarity
among sentences to mine positive pairs of relevant video moments from different videos as well as negatives pairs for contrastive learning.

Existing weak-supervised methods ignore the influence of intra-sample confrontment between semantically similar moments, they hence fail to distinguish the target moment from plausible negative moments. Inspired by this, Zhang et al.~\cite{RTBPN} advanced a Regularized Two-Branch Proposal Network (RTBPN) that simultaneously considers the inter-sample and intra-sample confrontments. 
To well model the fine-grained video-text local
correspondences, Yang et al.~\cite{lcnet} presented a Local Correspondence Network (LCNet) for weakly supervised temporal
sentence grounding. Specifically, it leverages the hierarchical feature representation module to model the one-one, one-many, many-one, and many-many correspondences between the video and text. 
Differently, Wang et al.~\cite{wstan} treated the weakly supervised task as a multiple instance learning (MIL) problem, and introduced a  Weakly Supervised Temporal
Adjacent Network (WSTAN) by integrating the temporal adjacent network, a complementary branch, and the self-discriminating loss into a unified framework. Therein,
the temporal adjacent network is leveraged to model the relationships among candidate proposals; the complementary branch is utilized to refine the predictions and rediscover more semantically meaningful clips; and the self-discriminating loss is designed to force the model to be more temporally discriminative. Similarly, 
Li et al.~\cite{ms2d} proposed a Multi-Scale 2D Representation Learning (MS-2D-RL) method, which  conducts convolution over multi-scale 2D temporal maps to capture the moment relations. Besides, it designs a  moment evaluation module to generate pseudo label for training.

\begin{table}[t]
  \centering
  \caption{Statistics of datasets for video moment localization task.}
  \vspace{-2ex}
  \label{Dataset}
  \small
  \scalebox{1.0}{\begin{tabular}{*{6}{c}}
    \hline
    Dataset & \#Videos & \#Queries & Duration & Domain & Source \\
    \hline
    DiDeMo~\cite{Lisa2017Localizing} & 10,464 & 40,543 & 30s & Open & Flickr \\
    TEMPO~\cite{Lisa2018Localizing} & 10,464 & - & 30s   & Open & Flickr \\
    Charades-STA~\cite{Jiyang2017TALL} & 9,848 & 16,128 & 31s  & Daily activities & Homes \\
    TACoS~\cite{Michaela2013Grounding} & 127 & 18,818 & 296s  & Cooking & Lab Kitchen \\
    ActivityNet-Captions~\cite{Ranjay2017Dense} & 19,209 & 71,957 & 180s & Open & YouTube \\
    \hline
  \end{tabular}}
\end{table}
\begin{table}[t]
  \centering
  \caption{Performance comparison between two supervised localization models on TEMPO-HL.}
  \vspace{-2ex}
  \label{TEMPO-HL}
  \small
  \scalebox{0.8}{\begin{tabular}{*{15}{c}}
    \hline
    \multirow{2}{*}{Methods} &\multirow{2}{*}{Year}& \multicolumn{2}{c}{DiDeMo} & \multicolumn{2}{c}{Before} & \multicolumn{2}{c}{After} & \multicolumn{2}{c}{Then} & \multicolumn{2}{c}{While} & \multicolumn{2}{c}{Average} \\
    \cmidrule{3-15}
     && R@1 & mIoU & R@1 & mIoU & R@1 & mIoU & R@1 & mIoU & R@1 & mIoU & R@1 & R@5 & mIoU \\
    \midrule
 MLLC~~\cite{Lisa2018Localizing}&2018 & 
 27.38& \textbf{42.45} &32.33& 56.91 &14.43& 37.33 &19.58& 50.39 &10.39& 35.95& 20.82& 71.68& 44.57\\
TCMN~~\cite{Songyang2019Exploiting} &2019& \textbf{28.77} & 42.37 & \textbf{35.47} & \textbf{59.28} & 17.91 & \textbf{40.79} & 20.47 & \textbf{50.78} & \textbf{18.81} & \textbf{42.95} & \textbf{24.29} & \textbf{76.98} & \textbf{47.24} \\
\bottomrule
\end{tabular}}
\end{table}
\begin{table}[t]
  \centering
  \caption{Performance comparison between two supervised localization models on TEMPO-TL.}
  \vspace{-2ex}
  \label{TEMPO-TL}
  \small
  \scalebox{0.8}{\begin{tabular}{*{13}{c}}
    \hline
    \multirow{2}{*}{Methods} &\multirow{2}{*}{Year}& \multicolumn{2}{c}{DiDeMo}&\multicolumn{2}{c}{Before} & \multicolumn{2}{c}{After} & \multicolumn{2}{c}{Then} & \multicolumn{2}{c}{Average} \\
    \cmidrule{3-13}
     & &R@1 & mIoU & R@1 & mIoU & R@1 & mIoU & R@1 & mIoU & R@1 & R@5 & mIoU \\
    \midrule
MLLC~~\cite{Lisa2018Localizing} & 2018&27.46 & \textbf{41.20} & 35.31 & 41.81 & 29.38 & 38.90 & 26.83 & 54.97 & 29.74 & 76.76 & 44.22 \\
TCMN~~\cite{Songyang2019Exploiting} & 2019&\textbf{28.90} & 41.03 & \textbf{37.68} & \textbf{44.78} & \textbf{32.61} & \textbf{42.77} & \textbf{31.16} & \textbf{55.46} & \textbf{32.85} & \textbf{78.73} & \textbf{46.01}\\
\bottomrule
\end{tabular}}
\vspace{-2ex}
\end{table}
\subsection{Reinforcement Learning}
To the best of our knowledge, Boundary Adaptive Refinement (BAR)~\cite{BAR} framework is the first work that extends reinforcement learning to weakly supervised video moment localization. Particularly, it consists of a context-aware feature extractor, an adaptive action planner, and a cross-modal alignment evaluator. Therein, the context-aware feature extractor is utilized to encode the current and contextualized environment states. The adaptive action planner is designed to infer action sequences to refine the temporal boundary, and the advantage actor-critic (A2C) algorithm
is chosen to train the adaptive action planner. Moreover, the cross-modal alignment evaluator is utilized to estimate the alignment score between each segment-query pair and assign a target-oriented reward to each action. 

\textbf{Summarization.} Among two-stage methods, TGA aims to learn a joint embedding space for video and query representations. Moreover, WSLLN and VLANet focus on proposal selection. To be specific, WSLLN leverages a two-stream structure to measure moment-query consistency and conduct moment selection simultaneously, while VLANet proposes a surrogate proposal selection module to prune out spurious moment candidates. Differently, LoGAN is devoted to learning contextualized visual representation. For one-stage methods, according to the processing of moment candidates, they are roughly divided into anchor-based and enumeration based approaches. Particularly, among anchor-based methods, SCN leverages the semantic completion module to refine the matching score, while VCA targets to learn more discriminative and robust visual features. As to enumeration based methods,  RTBPN aims to tackle the inter-sample and intra-sample confrontment issue. LCNet focuses on exploring one-one, one-many, many-one, and many-many interaction information between the video and text. WSTAN designs the temporal adjacent network to model the relationships among moment candidates, while MS-2D-RL conducts convolution over multi-scale 2D temporal maps to capture  moment relations.
\begin{table}[t]
  \centering
  \caption{Performance comparison among supervised localization models on DiDeMo.}
  \vspace{-2ex}
  \label{DiDeMo}
  \small
  \scalebox{1.0}{\begin{tabular}{*{6}{c}}
    \hline
    Type & Method &Year& R@1 & R@5 & mIoU \\
    \midrule
\multirow{3}{*}{Two-stage} & MCN~~\cite{Lisa2017Localizing}& 2017& 28.10 & 78.21 & 41.08 \\
&MLLC~~\cite{Lisa2018Localizing}&2018 & 28.37& 78.64 &\textbf{43.65}\\
& TCMN~~\cite{Songyang2019Exploiting} & 2019&\textbf{28.90} & \textbf{79.00} & 41.03 \\

\midrule                           
\multirow{5}{*}{One-stage} &
TGN~~\cite{Jingyuan2018Temporally} & 2018&28.23& 79.26 & 42.97 \\

&TMN~~\cite{Bingbin2018Temporal}&2018& 22.92 & 76.08 & 35.17 \\
& LNet~~\cite{Jingyuan2019Localizing} &2019&- &  -&  41.43 \\
& MAN~~\cite{Da2019MAN} &2019& 27.02 & \textbf{81.70} & 41.16 \\
&I2N~~\cite{iin}&2021&\textbf{29.00}&73.09&\textbf{44.32}\\
\midrule        
RL & SM-RL~~\cite{Weining2019Language} & 2019&31.06 & 80.45 & 43.94 \\
\bottomrule
\end{tabular}}
\vspace{-2ex}
\end{table}

\begin{table}[t]
  \centering
  \caption{Performance comparison among supervised localization models on TACoS.}
  \vspace{-2ex}
  \label{TACoS}
  \small
  \scalebox{0.9}{\begin{tabular}{ccccccccc}
    \toprule
     Type&Model&Year& R(1,0.5) & R(1,0.3) & R(1,0.1) & R(5,0.5) & R(5,0.3) & R(5,0.1)\\
    \midrule
\multirow{9}{*}{Two-stage} & CTRL~~\cite{Jiyang2017TALL}&2017& 13.30 & 18.32 & 24.32 & 25.42 & 36.69 & 48.73 \\

& ACRN~~\cite{Meng2018Attentive}&2018& 14.62 & 19.52 & 24.22 & 24.88 & 34.97 & 47.42 \\

& VAL~~\cite{Xiaomeng2018VAL} &2018&  14.74 & 19.76 & 25.74 & 26.52 & 38.55 & 51.87 \\
& MCF~~\cite{Aming2018Multi}&2018  & 12.53 & 18.64 & 25.84 & 24.73 & 37.13 & 52.96 \\
& SLTA~\cite{Bin2019Cross} &2019& 11.92 & 17.07 & 23.13 & 20.86 & 32.90 & 46.52 \\
& ACL~~\cite{Runzhou2019MAC} &2019& 20.01& 24.17 & 31.64 & 30.66 & 42.15 & 57.85 \\
& SAP~~\cite{Shaoxiang2019Semantic}& 2019&18.24 & - & 31.15 & 28.11 & - & 53.51 \\
&BPNet~~\cite{BPNet}&2021&20.96&25.96&-&-&-&-\\
&MMRG~~\cite{mmrg}&2021&\textbf{39.28}&\textbf{57.83}&\textbf{85.34}&\textbf{56.34}&\textbf{78.38}&\textbf{84.37}\\
\midrule                           
\multirow{33}{*}{One-stage} & TGN~~\cite{Jingyuan2018Temporally} & 2018& 18.90 & 21.77 & 41.87 & 31.02 & 39.06 & 53.40 \\
& BSSTL~~\cite{Cheng2019Bidirectional} & 2019&18.73 & 22.31 & 43.11 & 29.89 & 40.87 & 54.67 \\
& CMIN~~\cite{Zhu2019Cross} & 2019& 18.05 & 24.64 & 32.48 & 27.02 & 38.46 & 62.13 \\
& ABLR~~\cite{Yitian2019To} & 2019&9.40 & 19.50 & 34.70 & - & - & - \\
& ExCL~~\cite{Soham2019ExCL} & 2019&28.00 & 45.50 & -& -& -& -\\
& SCDM~~\cite{Yitian2019Semantic} & 2019&21.17 & 26.11 &-  & 32.18 & 40.16 & -\\
& DEBUG~~\cite{Chujie2019DEBUG} & 2019& - & 23.45 & 41.15 & - & - & - \\
& VSLNet~~\cite{Hao2020Span}  & 2020&24.27 & 29.61 & - & - & - & - \\
& DRN~~\cite{Runhao2020Dense} & 2020&23.17 & - & - & 33.36 & - & - \\
& CBP~~\cite{Jingwen2020The}& 2020&24.79 & 27.31 & - & 37.40 & 43.64 & -\\
& CMIN-R~~\cite{cmin_ext} & 2020& 19.57 & 27.33 & 36.88 & 28.53 & 43.35 & 64.93 \\
& CSMGAN~~\cite{Daizong2020Jointly} &2020& 27.09 & 33.90 & 42.74 & 41.22 & 53.98 & 68.97 \\
& FIAN~~\cite{Xiaoye2020Fine} & 2020&28.58 & 33.87 & 39.55 & 39.16 & 47.76 & 56.14 \\
& RMN~~\cite{liu-etal-2020-reasoning} & 2020&25.61 & 32.21& 42.17& 40.58 & 54.20 & 68.75 \\
& 2D-TAN~~\cite{Songyang2020Learning} &2020& 25.32 & 37.29 & 47.59 & 45.04 & 57.81 & 70.31 \\
&DPIN~~\cite{DIN}& 2020& 32.92&46.74&59.04&50.26&62.16&75.78\\
&CLEAR~~\cite{clear}&2021&30.27&42.18&-&51.76&63.61&-\\
&IA-Net~~\cite{ranet}&2021&26.27&37.91&47.18&46.39&57.62&71.75\\
&APGN~\cite{APGN}&2021&27.86&40.47&-&47.12&59.98&-\\

&CMHN~~\cite{cmhn}&2021&25.58&30.04&-&35.23&44.05&-\\

&SMIN~~\cite{smin}&2021&35.24&48.01&-&53.36&65.18&-\\
&RaNet~~\cite{ranet}&2021&33.54&43.34&-&55.09&67.33&-\\

&IVG~~\cite{IVG}&2021&29.07&38.84&49.36&-&-&-\\
&CPN~~\cite{cpn}&2021&36.58&48.29&\textbf{61.24}&-&-&-\\
&CBLN~~\cite{cbln}&2021&27.65&38.98&49.16&46.24&59.96&73.12\\
&MS-2D-TAN~~\cite{MS-2D-TAN}&2021&34.29&41.74&49.24&56.76&67.01&\textbf{78.33}\\
&I2N~\cite{iin}&2021&29.25&31.47&-&46.08&52.65&-\\
&FVMR~\cite{fvmr}&2021&29.12&41.48&53.12&50.00&64.53&78.12\\
&ACRM~\cite{acrm}&2021&33.84&47.11&-&-&-&-\\
&MIM~\cite{liang2021local}&2021&26.54&35.44&44.69&-&-&\\
&CP-Net~\cite{cp-net}&2021&29.29&42.61&-&-&-&-\\
&DORi~\cite{dori}&2021&28.69&31.80&-&-&-&-\\
&MATN~~\cite{matn}&2021&\textbf{37.57}&\textbf{48.79}&-&\textbf{57.91}&\textbf{67.63}&-\\
\midrule
\multirow{2}{*}{RL} 
& SM-RL~~\cite{Weining2019Language}&2019& 15.95 & 20.25 & 26.51 & 27.84 & 38.47 & 50.01 \\
& TripNet~~\cite{Meera2019Tripping}& 2020&\textbf{19.17} & \textbf{23.95} & -  &  - & -  &-   \\
\bottomrule
\end{tabular}}
\vspace{-2ex}
\end{table}
\section{Unsupervised Video Moment Localization}
Although weakly-supervised video moment localization approaches have achieved favorable performance, a certain amount of paired video-sentence data is still indispensable for model learning. Moreover, it is difficult to collect a large-scale video-sentence dataset for an arbitrary domain in the wild. Inspired by this, Gao et al.~\cite{u-vmr} proposed the unpaired video moment retrieval approach, namely U-VMR, which requires no textual annotations of videos and instead leverages the existing visual concept detectors and a pre-trained image-sentence embedding space. To be specific, it first designs a video-conditioned sentence generator to generate suitable sentence representations by leveraging the mined visual concepts. And then it develops a relation-aware moment localizer, which leverages a sentence-guided graph neural network to effectively select a portion of video clips as the moment representation. To obtain
different candidate moments, U-VMR follows 2D-TAN~\cite{Songyang2020Learning} to utilize a sparse sampling strategy for generating
candidate moments. Finally, the pre-trained image-sentence embedding model is adopted to align the generated sentence and moment representations for model learning.  Similarly, Nam et al.~\cite{PSVL} proposed the zero-shot natural language
video localization task, which only requires easily available text corpora, off-the-shelf object detector, and a collection of videos to localize. To address
the task, a Pseudo-Supervised Video Localization (PSVL) method is advanced. To be specific, it first generates temporal event proposal by compositing discovered atomic event, and then generates simplified sentence for each temporal event proposal via considering nouns detected by the object detector and verbs predicted from text corpora. Finally, it utilizes a attentive cross-modal neural network to predict the temporal boundary regions.

\begin{table}[t]
  \centering
  \caption{Performance comparison among supervised localization models on Charades-STA.}
  \vspace{-2ex}
  \label{Charades-STA}
  \small
  \scalebox{0.9}{\begin{tabular}{ccccccccc}
    \toprule
   Type& Model&Year&R(1,0.7) & R(1,0.5) & R(1,0.3) & R(5,0.7) & R(5,0.5) & R(5,0.3) \\
    \midrule
\multirow{10}{*}{Two-stage} & CTRL~~\cite{Jiyang2017TALL}& 2017&7.15 & 21.42 & - & 26.91 & 59.11 & -\\

&ROLE~~\cite{liumm2018}&2018&7.82&21.74&37.68&30.06&70.37&92.79\\
& VAL~~\cite{Xiaomeng2018VAL} &2018& 9.16 & 23.12 & - & 27.98 & 61.26 & - \\
& SLTA~~\cite{Bin2019Cross} & 2019&8.25 & 22.81 & 38.96 & 31.46 & 72.39 & 94.01 \\
& ACL~~\cite{Runzhou2019MAC}& 2019&12.20 & 30.48 & - & 35.13 & 64.84 & -\\

& QSPN~~\cite{Huijuan2019Multilevel}&2019& 15.80 & 35.60 & 54.70 & 45.40 & \textbf{79.40} & \textbf{95.60} \\
& SAP~~\cite{Shaoxiang2019Semantic} & 2019&13.36 & 27.42 & - & 38.15 & 66.37 & -\\
&MIGCN~~\cite{migcn}&2021&22.04&42.26&-&-&-&-\\
&BPNet~~\cite{BPNet}&2021&20.51&38.25&55.46&-&-&-\\
&MMRG~~\cite{mmrg}&2021&-&\textbf{44.25}&\textbf{71.60}&-&\textbf{60.22}&78.67\\
\midrule
\multirow{33}{*}{One-stage} 
& SCDM~~\cite{Yitian2019Semantic} &2019& 33.43 & 54.44 &- & 58.08 & 74.43 &- \\
& MAN~~\cite{Da2019MAN} &2019&22.72 & 46.53 & - & 53.72 & 86.23 &- \\
& DEBUG~~\cite{Chujie2019DEBUG} & 2019&17.69 & 37.39 & 54.95 & - & - & - \\
& ExCL~\cite{Soham2019ExCL}& 2019&22.40 & 44.10 & 61.50 &- &- &- \\
& CBP~~\cite{Jingwen2020The} & 2020&18.87 & 36.80 &- & 50.19 & 70.94 &- \\
& PMI-LOC~~\cite{Shaoxiang2020Learning} & 2020& 19.27 & 39.37 & 55.48 & - & - & - \\
& FIAN~~\cite{Xiaoye2020Fine} & 2020&37.72 & 58.55 & - & 63.52 & 87.80 & - \\
& RMN~~\cite{liu-etal-2020-reasoning} & 2020&36.98 & 59.13 & - & 61.02 & 87.51 & - \\
& 2D-TAN~~\cite{Songyang2020Learning} &2020& 23.25 & 39.81 & -& 52.15 & 79.33 & -\\
&DPIN~~\cite{DIN}&2020&26.96&47.98&-&55.00&85.53&-\\
& PFGA~~\cite{Cristian2020Proposal} &2020& 33.74 & 52.02 & 67.53 & -&- &- \\
& VSLNet~~\cite{Hao2020Span} & 2020&30.19 & 47.31 & 64.30 & - & - & - \\
& DRN~~\cite{Runhao2020Dense} & 2020&23.68 & 42.90 & - & 54.87 & 87.80 & - \\ 
&HVTG~~\cite{HTV}&2020&23.30& 47.27&61.37&-&-&-\\
&APGN~~\cite{APGN}&2021&38.86&62.58&-&62.11&91.24&-\\
&SMIN~~\cite{smin}&2021&\textbf{40.75}&\textbf{64.06}&-&\textbf{68.09}&89.49&-\\
&RaNet~~\cite{ranet}&2021&39.65&60.40&-&64.54&89.57&-\\
&IA-Net~~\cite{ranet}&2021&37.91&61.29&-&62.04&\textbf{89.78}&-\\
&SSMN~~\cite{SSMN}&2021&28.49&51.51&66.13&-&-&-\\
&IVG~~\cite{IVG}&2021&32.88&50.24&67.63&-&-&-\\
&CPN~~\cite{cpn}&2021&36.67&59.77&75.53&-&-&-\\
&CBLN~~\cite{cbln}&2021&28.22&47.94&-&57.47&88.20&-\\
&DRFT~\cite{drft}&2021&40.15&63.03&\textbf{76.68}&-&-&-\\
&LP-Net~\cite{LP-Net}&2021&34.03&54.33&66.59&-&-&-\\
&MQEI~\cite{mqei}&2021&38.04&59.35&73.67&-&-&-\\
&MS-2D-TAN~\cite{MS-2D-TAN}&2021&23.25&41.10&-&48.55&81.53&-\\
&I2N~\cite{iin}&2021&22.88&41.69&-&46.85&75.73&-\\
&FVMR~\cite{fvmr}&2021&18.22&38.16&-&44.96&82.18&-\\
&ACRM~\cite{acrm}&2021&22.60&44.10&65.10&-&-&-\\
&SV-VMR~\cite{sv-vmr}&2021&19.98&38.09&-&38.15&66.37&-\\
&MIM~\cite{liang2021local}&2021&35.51&53.23&70.08&-&-&-\\
&CP-Net~\cite{cp-net}&2021&22.47&40.32&-&-&-&-\\
&DORi~\cite{dori}&2021&40.56&59.65&72.72&-&-&-\\
\midrule
\multirow{5}{*}{RL} 
& RWM~~\cite{Dongliang2019Read} &2019& -& 36.70 &-  & - &  -&-  \\
& SM-RL~~\cite{Weining2019Language}&2019&- & 11.17 & 24.36 &- & 32.08 & 61.25 \\
& TSP-PRL~~\cite{Jie2020Tree} & 2020& 24.75 & 45.45 & - & - & - & - \\
& TripNet~~\cite{Meera2019Tripping}& 2020& 16.07 & 38.29 & \textbf{54.64} &-  & - & - \\
& MABAN~~\cite{maban}&2021&\textbf{32.26}&\textbf{56.29}&-&-&-&-\\
\bottomrule
\end{tabular}}
\vspace{-4ex}
\end{table}

\begin{table}[t]
  \centering
  \caption{Performance comparison among supervised localization models on ActivityNet-Captions.}
  \vspace{-2ex}
  \label{activity}
  \small
  \scalebox{0.9}{\begin{tabular}{ccccccccc}
    \toprule
   Type&Model&Year&R(1,0.7) & R(1,0.5) & R(1,0.3) & R(5,0.7) & R(5,0.5) & R(5,0.3) \\
    \midrule
\multirow{3}{*}{Two-stage} 

& QSPN~~\cite{Huijuan2019Multilevel} & 2019&13.60 & 27.70 & 45.30 & 38.30 & 59.20 & 75.70 \\
&MIGCN~~\cite{migcn}&2021&\textbf{44.94}&\textbf{60.03}&-&-&-&-\\
&BPNet~~\cite{BPNet}&2021&24.69&42.07&\textbf{58.98}&-&-&-\\
\midrule
\multirow{39}{*}{One-stage} & TGN~~\cite{Jingyuan2018Temporally}&2018&-  & 28.47 & 45.51 & - & 43.33 & 57.32 \\
& BSSTL~~\cite{Cheng2019Bidirectional} &2019& - & 47.68 & 55.32 & - & 57.53 & 70.53 \\
& CMIN~~\cite{Zhu2019Cross}& 2019&23.88 & 43.40 & 63.61 & 50.73 & 67.95 & 80.54 \\
& SCDM~~\cite{Yitian2019Semantic} &2019& 19.86 & 36.75 & 54.80 & 41.53 & 64.99 & 77.29 \\
& ABLR~~\cite{Yitian2019To}& 2019&- & 36.79  & 55.67 & - & - &- \\
& ExCL~~\cite{Soham2019ExCL} & 2019&23.60 & 43.60 & 63.00 & - & - & - \\
& DEBUG~~\cite{Chujie2019DEBUG} &2019& - & 39.72 & 55.91 & - & - & - \\
& CBP~~\cite{Jingwen2020The} & 2020&17.80 & 35.76 & 54.30 & 46.20 & 65.89 & 77.63 \\
& CMIN-R~~\cite{cmin_ext}& 2020&24.48 & 44.62 & 64.41 & 52.96 & 69.66 & 82.39 \\
& PMI-LOC~~\cite{Shaoxiang2020Learning} &2020& 17.83 & 38.28 & 59.69 & - & - & - \\
& CSMGAN~~\cite{Daizong2020Jointly} &2020& 29.15 & \textbf{49.11} & \textbf{68.52} & 59.63 & 77.43 & 87.68 \\
& FIAN~~\cite{Xiaoye2020Fine} & 2020&29.81 & 47.90 & 64.10 & 59.66 & 77.64 & 87.59 \\
& RMN~~\cite{liu-etal-2020-reasoning} & 2020&27.21 & 47.41 & 67.01 & 56.76 & 75.64 & 87.03 \\
& 2D-TAN~~\cite{Songyang2020Learning}  &2020& 27.38  & 44.05 & 58.75 & 62.26 & 76.65 & 85.65 \\
&DPIN~~\cite{DIN}&2020&28.31&47.27&62.40&60.03&77.45&87.52\\
& PFGA~~\cite{Cristian2020Proposal} & 2020&19.26 & 33.04 & 51.28 &-  & - & - \\
& VSLNet~~\cite{Hao2020Span}  & 2020&26.16 & 43.22 & 63.16 & - & - & - \\
& SQAN~~\cite{Jonghwan2020Local} & 2020&23.07 & 41.51 & 58.52 & - & - & - \\
& DRN~~\cite{Runhao2020Dense}  & 2020&24.36 & 45.45 & - & 50.30 & 77.97 & - \\
&HVTG~~\cite{HTV}&2020&18.27&40.15&57.60&-&-&-\\
&APGN~\cite{APGN}&2021&28.64&48.92&-&63.19&78.87&-\\
&CLEAR~~\cite{clear}&2021&28.05&45.33&59.96&62.13&77.26&85.83\\
&IA-Net~~\cite{ranet}&2021&27.95&48.57&67.14&63.12&78.99&87.21\\
&SMIN~~\cite{smin}&2021&30.34&48.46&-&62.11&\textbf{81.16}&\\
&RaNet~~\cite{ranet}&2021&28.67&45.59&62.97&\textbf{75.93}&-\\
&MATN~~\cite{matn}&2021&\textbf{31.78}&48.02&-&63.18&78.02&-\\
&SSMN~~\cite{SSMN}&2021&20.03&35.38&52.76&-&-&-\\
&IVG~~\cite{IVG}&2021&27.10&43.84&63.22&-&-&-\\
&CPN~~\cite{cpn}&2021&28.10&45.10&62.81&-&-&-\\
&CBLN~~\cite{cbln}&2021&27.60&48.12&66.34&63.41&79.32&\textbf{88.91}\\
&DRFT~\cite{drft}&2021&27.79&45.72&62.91&-&-&-\\
&LP-Net~\cite{LP-Net}&2021&25.39&45.92&64.29&-&-&-\\
&MQEI~\cite{mqei}&2021&24.58&45.86&64.39&-&-&-\\
&MS-2D-TAN~\cite{MS-2D-TAN}&2021&29.21&46.16&61.04&60.85&78.80&87.30\\
&FVMR~\cite{fvmr}&2021&26.85&45.00&60.63&61.04&77.42&86.11\\
&CMHN~\cite{cmhn}&2021&24.02&43.47&62.49&53.16&73.42&85.37\\
&SV-VMR~\cite{sv-vmr}&2021&27.32&45.21&61.39&63.44&77.10&85.98\\
&CP-Net~\cite{cp-net}&2021&21.63&40.56&-&-&-&-\\
&DORi~\cite{dori}&2021&26.41&41.35&57.89&-&-&-\\
\midrule
\multirow{4}{*}{RL}
& RWM~~\cite{Dongliang2019Read}& 2019&- & 36.90 &-  & - &-  & - \\
& TSP-PRL~~\cite{Jie2020Tree} &2020& - & 38.82 & \textbf{56.02} & - & - & - \\
& TripNet~~\cite{Meera2019Tripping}& 2020&13.93 & 32.19 & 48.42 & - & - &  -\\
& MABAN~~\cite{maban}&2021&\textbf{23.05}&\textbf{40.01}&-&-&-&-\\
\bottomrule
\end{tabular}}
\vspace{-2ex}
\end{table}

\section{Datasets and Evaluation}
\subsection{Datasets}
The statistics of the datasets for video moment localization task are summarized in Table~\ref{Dataset}. 

\textbf{DiDeMo~\cite{Lisa2017Localizing}.} The Distinct Describable Moments (DiDeMo) dataset is recently proposed in~\cite{Lisa2017Localizing}. It contains 10,464 videos with 40,543 annotated queries. To annotate moment-query pairs, videos are trimmed to a maximum of 30 seconds and then divided into 6 segments with 5 seconds long each. Besides, each moment in this dataset is constructed by one or more consecutive segments. Therefore, there are 21 moment candidates in each video, and the task is shifted to selecting the moment that best matches the query. 

\textbf{TEMPO~\cite{Lisa2018Localizing}.} The TEMPO dataset is collected based on the DiDeMo dataset~\cite{Lisa2017Localizing}. Specifically, it further extends the language descriptions that involve multiple events, while keeping its videos the same. The extended language expressions are collected based on four commonly used temporal words, before, after, while, and then. Simple sentences that come from DiDeMo are also included in this dataset. There are two parts in TEMPO, i.e., TEMPO-TL (Template Language) which is constructed by the original DiDeMo sentences with language templates and TEMPO-HL (Human Language) which is built by human annotations. Note that since the download link of this dataset is invalid currently, we hence cannot obtain the statistic result of the queries.

\textbf{Charades-STA~\cite{Jiyang2017TALL}.} The Charades dataset~\cite{Gunnar2016Hollywood} is first collected from daily indoor activities for activity understanding. Each video contains temporal activity annotation (from 157 activity categories) and multiple video-level descriptions. To make it suitable for language-based temporal location task, Gao et al.~\cite{Jiyang2017TALL} decomposed the original video-level descriptions into shorter sub-sentences, and performed keyword matching to assign them to temporal segments in videos. The alignment annotations are further verified manually. In total, there are 9, 848 videos with 16,128 annotated queries in this dataset, and these videos are 31 seconds long on average.

\textbf{TACoS~\cite{Michaela2013Grounding}.} This dataset is constructed by~\cite{tacos}. It was built on the top of MPII Compositive dataset~\cite{Marcus2012Script}, which contains different activities in the cooking domain. In TACoS, each video is associated with two type of annotations. The first one is fine-grained activity labels with temporal location (start and end time). The second is natural language descriptions with temporal locations. Note that the natural language descriptions are obtained by crowd-sourcing annotators, who are asked to describe the content of the video clips by sentences. In total, there are 127 videos picturing people who perform cooking tasks with 18,818 queries. 

\textbf{ActivityNet-Captions~\cite{Ranjay2017Dense}.} This dataset is proposed by Krishna et al.~\cite{Ranjay2017Dense} for the dense video captioning task, which contains annotations from the open domain. In this dataset, each video contains at least two ground truth segments, and each segment is paired with one ground truth caption. In total, this dataset contains 19, 209 videos and 71,957 queries.

\subsection{Evaluation Metrics}
In existing studies, ``R@$n$, IoU=$m$'' proposed by~\cite{Hu_2016_CVPR} is commonly adopted as the evaluation metric to measure their performance. 
It is defined as the percentage of at least one of the top-$n$
predicted moments which have IoU with ground-truth moment larger than $m$~\cite{fvmr}. In the following, we use $R(n,m)$ to denote ``R@n, IoU=m''. Note that for DiDeMo and TEMPO datasets, existing methods measure their performance with Rank@1 (R@1), Rank@5 (R@5), and mean intersection over union (mIoU).

\section{Experimental Results}

\subsection{Supervised Localization Models}
Supervised video moment localization models are evaluated on the DiDeMo, TEMPO, TACoS, Charades-STA, and ActivityNet-Captions datasets. We directly summarized their experimental results from the corresponding papers in Tables~\ref{TEMPO-HL}-\ref{activity}.

\subsubsection{Two-stage ones} Among two-stage supervised video moment localization models, the hand-crafted heuristics based ones (i.e., MCN, MLLC, and TCMN) are mainly evaluated on DiDeMo and TEMPO. Their experimental results are reported in Table~\ref{TEMPO-HL}, \ref{TEMPO-TL}, and \ref{DiDeMo}. As one of the earliest introduced methods, MCN is considered the de facto baseline result. MLLC outperforms the baseline model MCN, suggesting that learning to reason about which context moment is correct (as opposed to considering  global video context) is beneficial. 
TCMN exhibits the promising performance across all the metrics of the complex sentence and comparative results in simple sentences on Tempo-HL. The results show that the compositional modeling of complex queries can improve the localization performance.

As to video moment localization approaches that utilize multi-scale sliding windows to generate moment candidates (i.e., CTRL, ACRN, VAL, SLTA, MMRG, ROLE, MCF, ACL, and MIGCN), experiments are primarily executed on TACoS and Charades-STA. The experimental results are reported in Table~\ref{TACoS} and \ref{Charades-STA}. We can see that: \textbf{1)} ROLE achieves better performance as compared to CTRL on Charades-STA, verifying the importance of visual-aware query modeling. \textbf{2)} Although ACRN, VAL, SLTA, and MMRG all focus on improving visual modeling, MMRG achieves superior performance on both TACoS and Charades-STA. This mainly because it jointly considers the object relations and the phrase relations modeling for video moment localization, meanwhile, it introduces pre-training tasks to enhance the
visual representation. \textbf{3)} Compared with MCF that also focus on intra-modal interaction modeling, ACL achieves remarkable performance improvements on both TACoS and Charades-STA. This reflects that the pair of activity concepts extracted from both videos and queries  play a vital role in improving the cross-modal alignment. \textbf{4)} MIGCN outperforms ACL on Charades-STA, justifying the necessity of capturing both intra- and inter-modal interaction information. And \textbf{5)} MMRG achieves superior performance, and the results are competitive to other multi-scale sliding windows based methods. One possible reason is that it adopts two self-supervised pre-training tasks: attribute masking and context prediction to alleviate semantic gap across modalities.

In contrast, the methods that leverages moment generation networks are mainly evaluated on Charades-STA and ActivityNet-Captions. As reported in  Table~\ref{Charades-STA} and \ref{activity}, BPNet outperforms both QSPN and SAP. Because it utilizes VSLNet to generate high-quality moment candidates, therefore boosting the localization accuracy. However, the localization accuracy of BPNet is significantly worse than the MMRG. The reasons may be that 1) Compared with VSLNet, the multi-scale sliding windows generate higher quality moment candidates. And 2) MMRG could identify the fine-grained differences among similar video moment candidates. Despite achieving promising performance, these two-stage approaches should pre-process the untrimmed videos to obtain moment candidates, therefore they are inferior in efficiency.

\subsubsection{One-stage ones} We summarized experimental results of all discussed one-stage methods in Table~\ref{DiDeMo}-\ref{activity}. From these results, we could find that: \textbf{1)} Among anchor-based one-stage methods, FIAN achieves the best performance on all datasets.  Particular, the performance of FIAN significantly surpasses TGN, demonstrating the importance of inter-modal interaction modeling. Moreover, compared with other inter-modal interaction modeling methods, FIAN achieves better performance. This verifies the effectiveness of the iterative attention mechanism. More importantly, FIAN even achieves superior performance as compared to methods that jointly consider intra- and inter-modal interaction modeling. This may be because it iteratively captures bilateral query-video interaction information. Furthermore, CMIN-R performs better than CMIN, verifying that reconstructing the natural language queries could indeed enhance the cross-modal representations. 
\textbf{2)} For sampler-based localization methods, SMIN achieves the best performance on  Charades-STA. However, MATN achieves the best performance on TACoS and ActivityNet-Captions and significantly surpasses the work SMIN. On one hand, this reflects that the enumeration strategy could generate higher quality moment candidates as compared to other strategies. On the other hand, it demonstrates the powerful ability of the visual-language transformer for learning modality representations. \textbf{3)} For proposal-free methods, CP-Net achieves the best performance on TACoS, but its performance is slightly worse on the other two datasets. This is mainly because the video length of TACoS is longer than the other two datasets, namely there are lots of visual similar moment candidates. CP-Net focuses on exploiting the fine-grained temporal clues to enhance the discriminative of different moments, therefore achieving the better performance on TACoS. As DORi could capture complex relationships between humans, objects and activities in the video, it outperforms CP-Net on Charades-STA which is collected for activity understanding. Compared to TACoS and Charades-STA datasets, ActivityNet-Captions is collected from Youtube, of which the videos contain multi-modal information. Thereby, DRFT achieves the best performance on  ActivityNet-Captions since  it utilizes multimodal information and adequately exploit interactions between modalities. And \textbf{4)} among all one-stage methods, MATN achieves superior performance as compared to FIAN, CP-Net, DRFT, and DORi. This reflects that the performance gap between moment generation based methods and proposal-free methods is still large.  Thereby, for proposal-free methods, it is worth to explore more effective interaction strategies to further improve localization accuracy.

\subsubsection{Reinforcement learning} Reinforcement learning based localization models alleviate the efficiency issue to a certain extent, yet their performance is inferior as reported in Table~\ref{DiDeMo}-\ref{activity}. The main reason may be that they mostly focus on the design of policy and rewards, ignoring the importance of multiple crucial factors, such as query representation learning, video context modeling, and multimodal fusion. 
\begin{table}[t]
  \centering
  \caption{Performance Comparison among weakly supervised localization models on DiDeMo.}
  \vspace{-2ex}
  \label{DiDeMo(weakly)}
  \small
  \scalebox{1.0}{\begin{tabular}{*{6}{c}}
    \hline
   Type &Model & Year&R@1 & R@5 & mIoU \\
    \midrule
\multirow{4}{*}{Two-stage} & TGA~~\cite{Niluthpol2019Weakly} &2019& 12.19 & 39.74 & 24.92 \\
& WSLLN~~\cite{Mingfei2019Wslln} &2019& 18.40 & 54.40 & 27.40 \\

& VLANet~~\cite{Minuk2020VLANet} &2020 & 19.32& \textbf{65.68} & 25.33 \\
&LoGAN~~\cite{logan}&2021&\textbf{39.20}&64.04&\textbf{38.28}\\
\midrule
\multirow{2}{*}{One-stage}&RTBPN~~\cite{RTBPN}&2020& \textbf{20.79}&\textbf{60.26}&29.81\\
&WSTAN~~\cite{wstan}&2021&19.40&54.64&\textbf{31.94}\\
    \bottomrule
  \end{tabular}}
  \vspace{-2ex}
\end{table}

\begin{table}[t]
  \centering
  \caption{Performance comparison among weakly supervised and unsupervised localization models on Charades-STA.}
  \vspace{-2ex}
  \label{Charades-STA(weakly)}
  \small
  \scalebox{0.9}{\begin{tabular}{ccccccccc}
    \toprule
     Type&Model&Year&R(1,0.7) & R(1,0.5) & R(1,0.3) & R(5,0.7) & R(5,0.5) & R(5,0.3) \\
   \midrule
\multirow{3}{*}{Two-stage} & TGA~~\cite{Niluthpol2019Weakly} &2019 & 8.84 & 19.94 & 32.14 & 33.51 & 65.52 & 86.58 \\
& VLANet~~\cite{Minuk2020VLANet} & 2020& 14.17& 31.83 & 45.24 & 33.09 & \textbf{82.85} & \textbf{95.70} \\
&LoGAN~~\cite{logan}&2021&\textbf{14.54}&\textbf{34.68}&\textbf{51.67}&\textbf{39.11}&74.30&92.74\\
\midrule
\multirow{6}{*}{One-stage}& SCN~~\cite{Zhijie2020Weakly} &2020& 9.97 & 23.58 & 42.96 & 38.87 & 71.80 & 95.56 \\
& RTBPN~~\cite{RTBPN} & 2020 & 13.24 & 32.36 & \textbf{60.04} & 41.18 & 71.85 & 97.48 \\
&LCNet~~\cite{lcnet}&2021&18.87&\textbf{39.19}&59.60&\textbf{45.24}&\textbf{80.56}&94.78\\
&VCA~~\cite{vca}&2021&\textbf{19.57}&38.13&58.58&37.75&78.75&\textbf{98.08}\\
&WSTAN~~\cite{wstan}&2021&12.28&29.35&43.39&41.53&76.13&93.04\\
&MS-2D-RL~~\cite{ms2d}&2021&17.31&30.38&-&34.92&69.60&-\\
\midrule
RL& BAR~~\cite{BAR}& 2020& 12.23& 27.04&44.97&-&-&-\\
\midrule
\multirow{2}{*}{Unsupervised} &U-VMR~~\cite{u-vmr}&2021&8.27&20.14&\textbf{46.69}&32.45&72.07&91.18\\
&PSVL~~\cite{PSVL}&2021&\textbf{14.17}&\textbf{31.29}&46.47&-&-&-\\
\bottomrule
\end{tabular}}
\vspace{-2ex}
\end{table}

\begin{table}[t!]
  \centering
  \caption{Performance comparison among weakly supervised and unsupervised localization models on  ActivityNet-Captions.}
  \vspace{-2ex}
  \label{activity weakly}
  \small
  \scalebox{0.9}{\begin{tabular}{ccccccccc}
    \toprule
    Type&Model&Year&R(1,0.5) & R(1,0.3) & R(1,0.1) & R(5,0.5) & R(5,0.3) & R(5,0.1) \\
    \midrule
\multirow{1}{*}{Two-stage} 

& WSLLN~~\cite{Mingfei2019Wslln} &2019& 22.70 & 42.80 & 75.40 & - & -&- \\

\midrule

\multirow{6}{*}{One-stage} & SCN~~\cite{Zhijie2020Weakly} &2020 & 29.22 & 47.23 & 71.48 & 55.69 & 71.45 & 90.88 \\
& RTBPN~~\cite{RTBPN} & 2020 & - & 29.63 & 49.77 & - & 60.59 & 79.89 \\
&LCNet~~\cite{lcnet}&2021&26.33&48.49&78.58&62.66&\textbf{82.51}&93.95\\
&VCA~~\cite{vca}&2021&\textbf{31.00}&50.45&67.96&53.83&71.79&92.14\\
&WSTAN~~\cite{wstan}&2021&30.01&\textbf{52.45}&\textbf{79.78}&\textbf{63.42}&79.38&\textbf{93.15}\\
&MS-2D-RL~\cite{ms2d}&2021&-&29.68&49.79&-&58.66&72.57\\
\midrule
RL&BAR~\cite{BAR}&2020&30.73&49.03&-&-&-&-\\
\midrule
\multirow{2}{*}{Unsupervised}&U-VMR~~\cite{u-vmr}&2021&11.64&26.38&\textbf{46.15}&30.83&54.27&73.13\\
&PSVL~~\cite{PSVL}&2021&\textbf{14.74}&\textbf{30.08}&44.74&-&-&-\\
\bottomrule
 \end{tabular}}
 \vspace{-3ex}
\end{table}

\subsection{Weakly Supervised and Unsupervised Localization Models}
Weakly supervised video moment localization approaches are mainly evaluated on the DiDeMo, Charadea-STA, and ActivityNet-Captions datasets. We directly summarized their experimental results from the corresponding papers in Table~\ref{DiDeMo(weakly)}, \ref{Charades-STA(weakly)}, and~\ref{activity weakly}.
TGA is the first weakly supervised video moment localization model, considered as the de facto baseline result. Among the two-stage approaches (i.e., WSLLN, LoGAN, and VLANet), LoGAN achieves the best performance on DiDeMo and Charadea-STA, as compared to WSLLN and VLANet. This reflects the importance of learning contextualized visual semantic representations for weakly-supervised moment localization. As to the one-stage models, VCA achieves the best performance in terms of ``R(1, 0.7)'' on both Charadea-STA and ActivityNet-Captions. This is mainly because enumeration based methods would generate overmuch moment candidates, while they cannot well learn discriminative and robust visual features under video-level supervision.  Differently, VCA utilizes the similarity among sentences to mine positive pairs as well as negatives pairs for contrastive learning, therefore learning more discriminative visual features.  
Besides, BAR as the first work that extends reinforcement learning to weakly supervised video moment localization also achieves promising performance. Particular, it outperforms TGA and SCAN on Charadea-STA and ActivityNet-Captions. 

It is noteworthy that unsupervised video moment localization approaches achieve favorable performance on both Charades-STA and ActivityNet-Captions. Specifically, PSVL obtains
comparable results against some supervised and weakly-supervised approaches on Charades-STA, such as ACL, SAP,  VSA-STV, RTBPN, and WSTAN. This verifies that even without any moment level or video-level annotations, we can also obtain acceptable localization results by carefully
designing the effective video-conditioned sentence generator.
\begin{table}[t!]
  \centering
  \caption{Efficiency comparison among some video localization models on three datasets. TE: time cost of query Embedding. CML: time cost of the cross-modal learning. ALL: the total time cost of TE and CML.}
  \vspace{-2ex}
  \label{efficiency}
  \small
  \scalebox{0.9}{\begin{tabular}{cccccccccc}
    \toprule
    \multirow{2}{*}{Methods} &\multicolumn{3}{c}{TACoS}&\multicolumn{3}{c}{ActivityNet Captions} & \multicolumn{3}{c}{Charades-STA}\\
    \cline{2-10}
    &TE&CML&ALL&TE&CML&ALL&TE&CML&ALL\\
    \toprule
PFGA~\cite{Cristian2020Proposal}&1.14&11.37&12.51&1.24&8.97&10.21&1.15&4.37&5.52\\
VSLNet~\cite{Hao2020Span}&3.58&5.02&8.59&3.87&4.86&8.74&3.90&4.27&8.18\\
SQAN~\cite{Jonghwan2020Local}&-&-&-&1.53&7.03&8.56&1.23&4.76&5.99\\
DRN~\cite{Runhao2020Dense}&4.67&22.13&26.81&4.86&18.46&23.32&4.52&12.39&16.91\\
CTRL~\cite{Jiyang2017TALL}& 4.32&534.23&538.55&4.75&398.25&403.00&4.53&12.20&16.73\\
SCDM~\cite{Yitian2019Semantic}&3.65&780.00&783.65&3.27&359.76&363.03&2.97&23.77&26.07\\
CBP~\cite{Jingwen2020The}&3.17&2659.01&2662.18&2.44&522.65&525.09&2.87&266.08&268.95\\
2D-TAN~\cite{Songyang2020Learning}&1.72&135.84&137.56&1.69&80.35&403.10&1.59&16.78&18.37\\
FVMR~\cite{fvmr}&3.51&0.14&3.65&3.14&0.09&3.23&2.86&0.01&2.87\\
\bottomrule
 \end{tabular}}
 \vspace{-4ex}
\end{table}

\subsection{Efficiency Comparison}
\cite{fvmr} conducts efficiency comparison among some video moment localization methods during inference, the results are summarized in Table~\ref{efficiency}. We can see that: 1) 
TE is not the test-time computational bottleneck since all methods cost similar time ($\sim$3ms). 2) CBP spends the most time in CML ($\sim$2600ms), demonstrating that self-attention technique as the contextual integration module for CML is time consuming. And 3)
FVMR is 35$\times$ to 20,000$\times$ faster than other methods, especially one-stage methods VSLNet and SQAN. This reflects that learning cross-modal common space is much more efficient than cross-modal interaction. Moreover, FVMR outperforms others in both speed and accuracy metrics, verifying that introducing common space learning strategy could well balance the efficiency and accuracy.
\section{DISCUSSION AND FUTURE DIRECTIONS}
The introduction of video moment localization task has aroused great interest, yet it is a challenging task since it requires joint reasoning over the visual and textual information. To better localize the target moment within the video, the video moment localization model should understand both the video and query information, such as objects, attributes, and actions, and then identify a particular moment via reasoning. Existing methods can be divided into supervised, weakly-supervised, and unsupervised learning paradigm based methods. Particular, the former two can be further classified into two-stage, one-stage, as well as reinforcement learning methods. We first discuss the main differences among them and point the possible technological trend of each type.  

Two-stage video moment localization approaches commonly utilize a separate scheme to generate moment candidates, and then match them with the query. Therefore, their localization efficiency is relatively low, and the localization accuracy is restricted by the moments generated by the first stage. \textit{In the future, it is a possible trend to design a more 
powerful moment generation model for the first-stage or a more efficient location regression module for the second-stage, to improve the localization efficiency and accuracy}. Compared with two-stage methods, 
one-stage ones do not need a separate stage for moment candidate generation, therefore they locate target moments faster. Some one-stage methods also utilize moment generation strategies to generate moment candidates, such as temporal convolution, but these processes are optimized together with the localization accuracy. This ensures the quality of generated moment candidates. However, existing models typically resort to various attention mechanisms to estimate the correlation between ``video-query'' pair, resulting in inefficiency and low scalability. \textit{In the future, building a effective and efficient cross-modal learning module, replacing the cross-modal interaction module, is deserved to exploring}. More importantly, both simple and complex input data are processed by the same interaction module, which is inappropriate. In light of this, \textit{building a novel interaction module, which could dynamically adopt different mechanisms to explore correlation information  for
different ``video-query'' pairs, is a possible technological trend}.  Different from the two- and one-stage methods, the reinforcement learning methods formulate this task as a problem of sequential decision making by learning an agent which regulates the temporal grounding
boundaries progressively based on its policy. In other words, they can localize more accurate temporal 
locations in a few glimpses of video.  Although they are more efficient, their localization accuracy is still  far from satisfactory due to the insufficient structural comprehension. \textit{One possible solution may be 
construct more efficient cross-modal interaction module to explore rich contextual information}.

Afterwards, we further point out several possible future directions to advance this research area. Current state-of-the-art models suffer serious dataset bias problem~\cite{otani2020challengesmr}. To be specific, on one hand, many queries describe the moments that appear at the start or end of the video. This imbalance makes the model can only predict the location of moments according to the statistical correlations without understanding the visual contents. On the other hand, as stated by Mithun et al.~\cite{Niluthpol2019Weakly}, it is difficult to mark the start and end locations of a certain moment. This may  introduce ambiguity in the training data and influence the alignment between visual and textual information. More importantly, sentence queries are typically simple and short, they mainly focus on one object and one action. In addition, many videos in existing datasets contain limited information, which require no complex reasoning. Therefore, \textit{building a large-scale dataset with long videos and semantic-rich queries would be an interesting research direction for this task}. Moreover, existing models tend to give a direct location prediction without an intermediate reasoning process. 
Therefore, it is difficult for people to evaluate the reasoning capability of models and analyze the localization results of the model. \textit{A promising research direction is to construct interpretable video moment localization models}.
Although recent approaches have achieved significant progress, their performance is still far below that of humans. This is because  humans have extensive domain knowledge. In other words, they can employ the corresponding background knowledge to successfully localize the target moment from the video, given the specific query. Thereby, \textit{it would be an interesting and promising direction to augment the video moment localization models with external knowledge}.

\section{Conclusion}
In this paper, we comprehensively review the state-of-the-art methods on video moment localization. To be more specific, we first review supervised learning based approaches, including two-stage, one-stage, and reinforcement learning models. 
We then describe the recently emerging weakly supervised and unsupervised video moment localization methods. After reviewing different video moment localization datasets, we group results according to the datasets. 
Finally, we figure out a number of promising directions for future research.

\begin{acks}

This work is supported by the National Natural Science Foundation of China, No.: U1936203 and No.: 62006142; the Shandong Provincial Natural Science Foundation for Distinguished Young Scholars, No.: ZR2021JQ26; the Major Basic Research Project of Natural Science Foundation of Shandong Province, No.: ZR2021ZD15; Science and Technology Innovation Program for Distinguished Young Scholars of Shandong Province Higher Education Institutions, No.: 2021KJ036; as well as the special fund for distinguished Professors of Shandong Jianzhu University.
\end{acks}

\bibliographystyle{ACM-Reference-Format}
\bibliography{sample-base}

\end{document}